\documentclass[10pt,twocolumn,letterpaper]{article}

\usepackage{iccv}
\usepackage{times}
\usepackage{epsfig}
\usepackage{graphicx}
\usepackage{amsmath}
\usepackage{amssymb}
\usepackage{csquotes}
\usepackage{mwe}
\usepackage{subcaption}
\usepackage[title]{appendix}
\usepackage{gensymb}
\usepackage{multirow}

\usepackage[breaklinks=true,bookmarks=false]{hyperref}

\iccvfinalcopy 

\def\iccvPaperID{5323} 

\ificcvfinal\fi

\begin{document}

\title{Physical Adversarial Textures That Fool Visual Object Tracking}

\author{Rey Reza Wiyatno \qquad Anqi Xu\\
Element AI\\
Montreal, Canada\\
{\tt\small \{rey.reza, ax\}@elementai.com}
}

\maketitle
\ificcvfinal\thispagestyle{empty}\fi



\begin{abstract}

We present a method for creating inconspicuous-looking textures that, when displayed as posters in the physical world, cause visual object tracking systems to become confused.
As a target being visually tracked moves in front of such a poster, its adversarial texture makes the tracker lock onto it, thus allowing the target to evade.
This adversarial attack evaluates several optimization strategies for fooling seldom-targeted regression models: non-targeted, targeted, and a newly-coined family of guided adversarial losses.
Also, while we use the Expectation Over Transformation (EOT) algorithm to generate physical adversaries that fool tracking models when imaged under diverse conditions, we compare the impacts of different scene variables 
to find practical attack setups with high resulting adversarial strength and convergence speed.
We further showcase that textures optimized using simulated scenes can confuse real-world tracking systems for cameras and robots.

\end{abstract}


\vspace{-0.4cm}
\section{Introduction} \label{sec:intro}

Research on adversarial attacks~\cite{42503,43405,DBLP:conf/eurosp/PapernotMJFCS16} have shown that deep learning models, e.g., for classification and detection tasks, are confused by adversarial examples: slightly-perturbed images of objects that cause them to make wrong predictions.
While early attacks digitally modified inputs to a victim model, later advances created photos~\cite{45818} and objects in the physical world that lead to misclassification under diverse imaging conditions~\cite{Eykholt_2018_CVPR,pmlr-v80-athalye18b}.
Due to these added complexities, many physical adversaries were not created to look \emph{indistinguishable} from regular items, but rather as \emph{inconspicuous} objects such as colorful eyeglasses~\cite{Sharif:2016:ACR:2976749.2978392,sharif2019general}.


\begin{figure}[h]
    \centering
    \begin{subfigure}{0.44\linewidth}
        \centering
        \includegraphics[width=\textwidth]{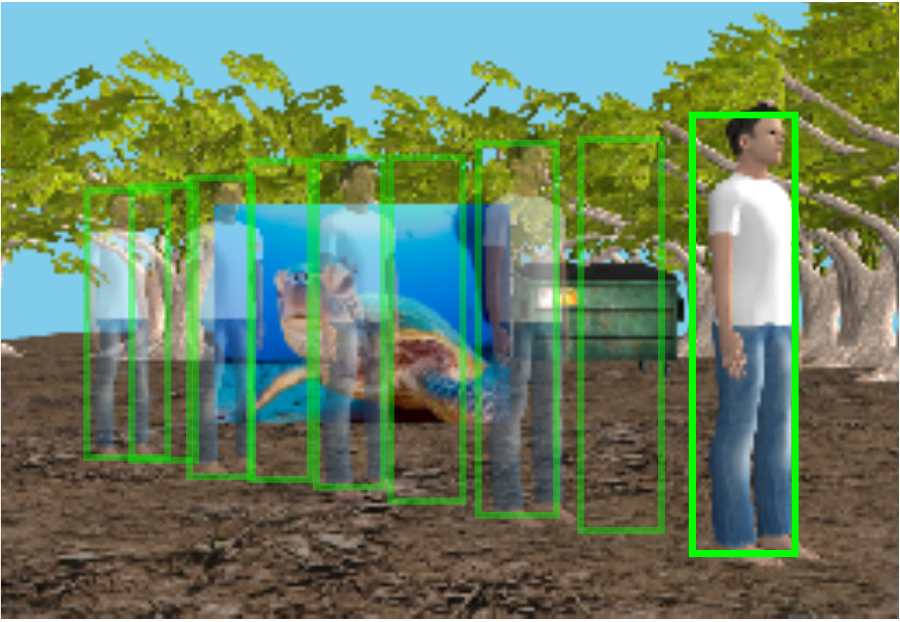}
        \caption{source texture}
        \label{fig:pat_sample_orig}
    \end{subfigure}
    \hbox{~}
    \begin{subfigure}{0.44\linewidth}
        \centering
        \includegraphics[width=\textwidth]{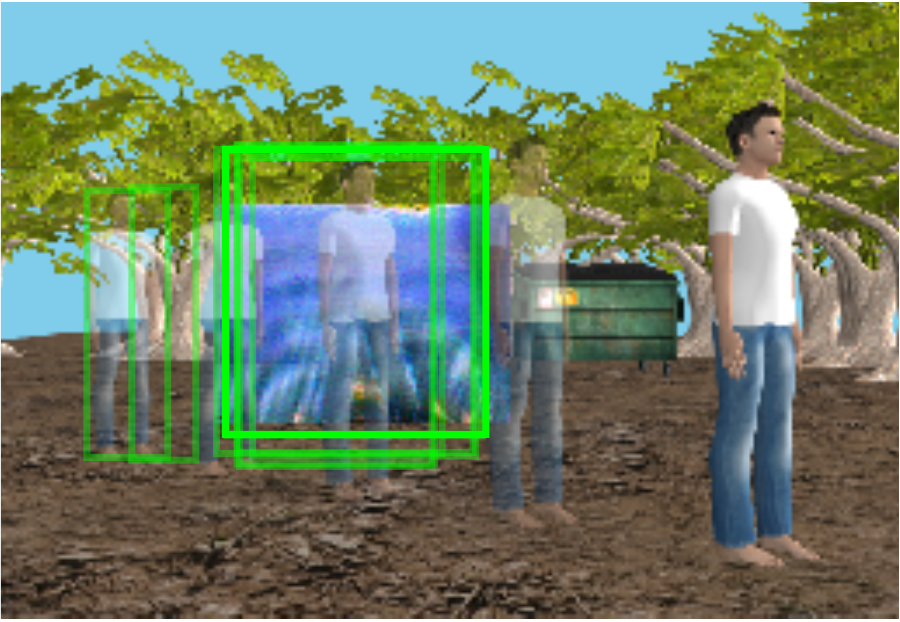}
        \caption{adversarial texture}
        \label{fig:pat_sample_adv}
    \end{subfigure}
    \caption{A poster of a Physical Adversarial Texture resembling a photograph, causes a tracker's bounding-box predictions to lose track as the target person moves over it.}
    \label{fig:pat_sample}
\vspace{-0.4cm}
\end{figure}

We study the creation of physical adversaries for an object tracking task, of which the goal is to find the bounding-box location of a target in the current camera frame given its location in the previous frame.
We present a method for generating Physical Adversarial Textures (PAT) that, when displayed as advertisement or art posters, cause regression-based neural tracking models like GOTURN~\cite{goturn2016} to break away from their tracked targets, even though these textures do not look like targets to human eyes, as seen in Figure~\ref{fig:pat_sample}.

Fooling a tracking system comes with added challenges compared to attacking classification or detection models.
Since a tracker adapts to changes in the target's appearance, an adversary must be universally effective as the target moves and turns.
Also, some trackers like GOTURN only search within a sub-region of the frame around the previous target location, and so only a small part of the PAT may be in view and not obstructed, yet it must still be potent.
Furthermore, it is insufficient for the tracker to be slightly off-target on any single frame, as it may still end up tracking the target semi-faithfully; robust adversaries must cause the system to break away from the tracked target over time.

Our main contributions are as follows:
\begin{enumerate}
    \item first known demo of adversaries for sequential tracking tasks, impacting domains such as surveillance, drone photography, and autonomous convoying,
    \item coining of \enquote{guided adversarial losses} concept, which strikes a middle-ground between targeted and non-targeted adversarial objectives, and empirically shown to enhance convergence and adversarial strength,
    \item study of Expectation Over Transformation (EOT)~\cite{pmlr-v80-athalye18b}, highlighting the need to randomize only certain scene variables while still creating potent adversaries, and
    \item show sim-to-real transfer of PATs created using a non-photorealistic simulator and diffuse-only materials.
\end{enumerate}


\section{Related Work} \label{sec:related}



Early \emph{white-box} physical adversarial attacks, which assumed access to the victim model's internals, created printable adversaries that were effective under somewhat varying views~\cite{45818}, by using gradient-based methods such as FGSM~\cite{43405}.
Similar approaches were employed to create eyeglass frames for fooling face recognition models~\cite{Sharif:2016:ACR:2976749.2978392,sharif2019general}, and
to make stop signs look like speed limits to a road sign classifier~\cite{Eykholt_2018_CVPR}.
Both latter systems only updated gradients within a \emph{mask}ed region in the image, namely over the eyeglass frame or road sign.
Still, neither work explicitly accounted for the effects of lighting on the imaged items.


Expectation Over Transformation (EOT)~\cite{pmlr-v80-athalye18b} formalized the strategy used by~\cite{Sharif:2016:ACR:2976749.2978392,Eykholt_2018_CVPR} of optimizing for adversarial attributes of a mask, by applying a combination of random transformations to it.
By varying the appearance and position of a 2-D photograph or 3-D textured object as the mask, EOT-based attacks~\cite{pmlr-v80-athalye18b,46561,liu2018beyond} generated physically-realizable adversaries that are robust within a range of viewing conditions.
Our attack also applies EOT, but we importantly study the efficacy and the need to randomize over different transformation variables, including foreground/background appearances, lighting, spatial locations of the camera, target, adversary, and surrounding objects.

CAMOU is a \emph{black-box} attack that also applied EOT to create adversarial textures for a car that made it non-detectable by object detection networks.
CAMOU approximated the gradient of an adversarial objective through both the complex rendering process and opaque victim network, by using a learned surrogate mapping~\cite{Papernot:2017:PBA:3052973.3053009} from the texture space directly onto the detector's confidence score.
Both their attack and evaluations were carried out using a photo-realistic rendering engine.
Still, this method was not tested in the real world, and also incurs high computational costs and potential instability risks due to the alternation optimizing the surrogate model and the adversarial perturbations.

DeepBillboard~\cite{DBLP:journals/corr/abs-1812-10812} attacked autonomous driving systems by creating adversarial billboards that caused a victim model to deviate its predicted steering angles within real-world drive-by sequences. While our work shares many commonalities with DeepBillboard, we confront added challenges by attacking a sequential tracking model rather than a per-frame regression network, and we also contrast the effectiveness of differing adversarial objectives.


\section{Object Tracking Networks} \label{sec:goturn}

Various learning-based tracking methods have been proposed, such as the recent GOTURN~\cite{goturn2016} deep neural network that regresses the location of an object in a camera frame given its previous location and appearance.
While other tracking methods based on feature-space cross-correlation~\cite{bertinetto2016fully,Valmadre_2017_CVPR} and tracking-by-detection~\cite{Feichtenhofer_2017_ICCV} are also viable, we focus on GOTURN models to ground our studies on the effectiveness of different types of adversarial losses, as well as the compute efficiency of an EOT-based attack.

As seen in Figure~\ref{fig:block_diagram}, given a target's bounding-box location $\hat{l}_{j-1}$ of size $w \times h$ in the previous frame $f_{j-1}$, GOTURN crops out the \emph{template} $\tilde{f}_{j-1}$ as a region of size $2 w \times 2 h$ around the target within $f_{j-1}$.
The current frame $f_{j}$ is also cropped to the same region, yielding the \emph{search area} $\tilde{f}_{j}$, which is assumed to contain most of the target still.
Both the template and search area are resized to $227 \times 227$ and processed through convolutional layers. 
The resulting feature maps are then concatenated and passed through fully-connected layers with non-linear activations, ultimately regressing $l_{j} = \{ (x_{min},y_{min}), (x_{max},y_{max}) \} \in [0,1]^4$, that is, the top-left and bottom-right coordinates of the target's location within the current search area $\tilde{f}_{j}$.


Such predictions can also be used for visual servoing, i.e., to control an aerial or wheeled robot to follow a target through space.
One approach~\cite{8287675,DBLP:conf/iros/ShkurtiC0IHLMXD17}
is to regulate the center-points and areas of predictions about the center of the camera frame and the desired target size, respectively, using Proportional-Integral-Derivative (PID) controllers on the forward/backward, lateral, and possibly vertical velocities of the vehicle.
In this work, we show that visual tracking models, as well as derived visual servoing controllers for aerial robots, can be compromised by PATs.


\section{Attacking Regression Networks} \label{sec:regression_attack}

For classification tasks, an adversarial example is defined as a slightly-perturbed version of a source image that satisfies two conditions: \emph{adversarial output} --- the victim model misclassifies the correct label, and \emph{perceptual similarity} --- the adversary is perceived by humans as similar to the source image.
We discuss necessary adjustments to both conditions when attacking regression tasks.
While recent work has shown the existence of adversaries that confuse regression tasks~\cite{NIPS2017_7273,DBLP:journals/corr/abs-1812-10812},
there is still a general lack of analysis on the strength and properties of adversaries as a function of different attack objectives.
In this work, we consider various ways to optimize for an adversary, and notably formalize a new family of \emph{guided} adversarial losses.
While this work focuses on images, the concepts discussed below are generally applicable to other domains as well, such as fooling audio transcriptions~\cite{NIPS2017_7273}.

\subsection{Adversarial Strength} \label{sec:adversarial_str}

There is no task-agnostic analog to misclassification for regression models, due to the non-discrete representation of their outputs.
Typically, a regression output is characterized as adversarial by thresholding a task-specific error metric.
This metric may also be used to quantify \emph{adversarial strength}.
For instance, adversaries for human pose-prediction can be quantified by the percentage of predicted joint poses beyond a certain distance from ground-truth locations~\cite{NIPS2017_7273}.
As another example, DeepBillboard~\cite{DBLP:journals/corr/abs-1812-10812} defines unsafe driving for an autonomous vehicle as experiencing an excessive amount of total lateral deviation, and quantifies adversarial strength as the percentage of frames in a given unit of time where the steering angle error exceeds a corresponding threshold.

When fooling a visual tracker, the end-goal is for the system to break away from the target \emph{over time}.
Therefore, we consider a sequence of frames $F^{\dagger} = \{f_{1}^{\dagger},f_{2}^{\dagger}, ...,f_{N}^{\dagger}\}$ where the target moves across a poster containing an adversarial texture $\chi$, and quantify adversarial strength by the average amount of overlap between tracker predictions $l_j$ (computed from $f_{j-1}^{\dagger}, f_{j}^{\dagger}$) and the target's actual locations $\hat{l_j}$.
We also separate the tracker's baseline performance from the effects of the adversary, by computing the average overlap ratio across another sequence $F = \{f_{1},f_{2}, ...,f_{N}\}$, in which the adversarial texture is replaced by an \emph{inert source texture}.
Thus, in this work, adversarial strength is defined by averaging the \textbf{mean-Intersection-Over-Union-difference} metric, $\mu IOUd$, over multiple generated sequences:

\small
\begin{align}
    IOU( l_j, \hat{l_j} ) =& \frac{ \mathcal{A}(l_j \cap \hat{l_j}) }{\mathcal{A}(l_j) + \mathcal{A}(\hat{l_j}) - \mathcal{A}(l_j \cap \hat{l_j})} \nonumber \\
    \mu IOUd =& \frac{1}{N-1} \sum_{j \in [2,N], f_j \in F} IOU\left( l_j(f_{j-1},f_j), \hat{l_j} \right) \label{eq:uioud} \\
    &- \frac{1}{N-1} \sum_{j \in [2,N], f_j^{\dagger} \in F^{\dagger}} IOU\left( l_j(f_{j-1}^{\dagger},f_j^{\dagger}), \hat{l_j} \right) \nonumber
\end{align}
\normalsize

\noindent where $\cap$ denotes the intersection of two bounding boxes and $\mathcal{A}(\cdot)$ denotes the area of the bounding box $l$.

\subsection{Perceptual Similarity} \label{sec:perceptual_sim}

Perceptual similarity is often measured by the distance between a source image and its perturbed variant, e.g., using Euclidean norm in the RGB colorspace~\cite{42503,DBLP:conf/sp/Carlini017}.
Sometimes, we apply a loose threshold to this constraint, to generate universal adversaries that remain potent under diverse conditions~\cite{Moosavi-Dezfooli_2017_CVPR,pmlr-v80-athalye18b,zhang2018camou}.
Other times, the goal is not to \emph{imitate} a source image, but merely to create an \emph{inconspicuous} texture that does not look harmful to humans, yet cause models to misbehave~\cite{Sharif:2016:ACR:2976749.2978392,46561,DBLP:journals/corr/abs-1812-10812}.
With this work, we aim to raise public awareness that \emph{colorful-looking art can be harmful to vision models}.

\subsection{Optimizing for Adversarial Behaviors} \label{sec:adversarial_loss}

While our attack's end-goal is to cause the tracker to break away from its target, we can encourage different \emph{adversarial behaviors}, such as locking onto part of an adversarial poster or focusing onto other parts of the scene.
These behaviors are commonly optimized into an adversary through loss minimization, e.g., using gradient descent.
The literature has proposed several families of adversarial losses, notably:

\begin{itemize}
    \item the baseline \textbf{non-targeted} loss $\mathcal{L}_{nt}$
    maximizes the victim model's training loss, thus causing it to become generally confused (e.g., FGSM~\cite{43405}, BIM~\cite{45818});
    
    \item \textbf{targeted} losses $\mathcal{L}_{t}$
    also apply the victim model's training loss, but to minimize the distance to an \emph{adversarial target output} (e.g., JSMA~\cite{DBLP:conf/eurosp/PapernotMJFCS16});
    
    \item we define \textbf{guided} losses $\mathcal{L}_{g}$ as middle-grounds between $\mathcal{L}_{nt}$ and $\mathcal{L}_{t}$, which regulate specific adversarial \emph{attributes} rather than strict output values, analogous to misclassification onto a set of output values~\cite{45818}; and
    
    \item \textbf{hybrid} losses use a weighted linear combination of the above losses to gain adversarial strength and speed up the attack (e.g., C\&W~\cite{DBLP:conf/sp/Carlini017}, Hot/Cold~\cite{DBLP:conf/cvpr/RozsaRB16} attacks).
\end{itemize}

The motivation for guided losses stems from our observations of the optimization rigidity of targeted losses, and weak guidance from the non-targeted loss.
Although similar ideas have been used~\cite{DBLP:conf/sp/Carlini017,DBLP:journals/corr/abs-1812-10812}, 
we formally coin \enquote{guided adversarial objectives} as those that regulate attributes of the victim model's output about specific adversarial values.

To fool object trackers, we consider these specific losses:

\begin{itemize}
    \item $\mathcal{L}_{nt} = -||l_j^{\dagger} - \hat{l_j}||_1$ increases GOTURN's training loss;
    \item $\mathcal{L}_{t-} = ||l_j^{\dagger} - \{(0.0,0.9), (0.1,1.0)\}||_1$ shrinks predictions towards the bottom-left corner of the search area;
    \item $\mathcal{L}_{t=} = ||l_j^{\dagger} - \{(0.25,0.25), (0.75,0.75)\}||_1$ predicts the exact location of the target in the previous frame;
    \item $\mathcal{L}_{t+} = ||l_j^{\dagger} - \{(0.0,0.0), (1.0,1.0)\}||_1$ grows predictions to the maximum size of the search area;
    \item $\mathcal{L}_{ga-} = \min( \mathcal{A}(l_j^{\dagger}) - \mathcal{A}(\hat{l_j}), 0 )$ encourages the area of each prediction to shrink from the ground-truth value;
    \item $\mathcal{L}_{ga+} = \max( \mathcal{A}(l_j^{\dagger}) - \mathcal{A}(\hat{l_j}), 0 )$: encourages the area of each prediction to grow from the ground-truth value.
\end{itemize}


Note that other guided losses are also possible, such as maximizing or minimizing the magnitudes of predictions.
For succinctness, we evaluated against a non-targeted loss and the simplest of targeted losses as baselines, to show that a well-engineered guided loss has the potential for better convergence and adversarial strength.

Additionally, we can enforce perceptual similarity by adding a Lagrangian-relaxed loss $\mathcal{L}_{ps}$~\cite{42503,DBLP:conf/sp/Carlini017,pmlr-v80-athalye18b}.
Its associated weight can be set heuristically, or fine-tuned via line search into the smallest value resulting in sufficient adversarial strength.
While most of our experiments generate \emph{inconspicuous} adversaries that do not enforce perceptual similarity, Section~\ref{sec:imitation} specifically showcases \emph{imitation} attacks.

In summary, our attack method optimizes a (possibly-imitated) source texture $\chi_0$ into an adversarial variant $\chi_i$ over $i \in [1,I_{max}]$ iterations, by minimizing a weighted linear combination of loss terms:

\begin{equation}
    \mathcal{L} = \bar{w} \cdot [\mathcal{L}_{nt}, \mathcal{L}_{t...}, \mathcal{L}_{g...}, \mathcal{L}_{ps}]^T \\
\end{equation}

\noindent
\noindent where the texture is incrementally updated as:

\begin{equation}
\chi_i = \chi_{i-1} + \alpha_i \cdot \Delta \chi
\end{equation}

\noindent Here, $\alpha_i$ denotes the step size at the $i$-th iteration, and $\Delta \chi$ denotes a perturbation term based on the gradient $\nabla_{\chi}\mathcal{L}$.



\section{Physical Adversarial Textures} \label{sec:method}

We now discuss how the above attack formulation can be generalized to produce Physical Adversarial Textures (PAT) that resemble colorful art.
Such PATs, when displayed on a digital poster and captured by camera frames near a tracked target, causes a victim model to lose track of the target.

\begin{figure*}[ht!]
\centering
\includegraphics[width=1.0\linewidth]{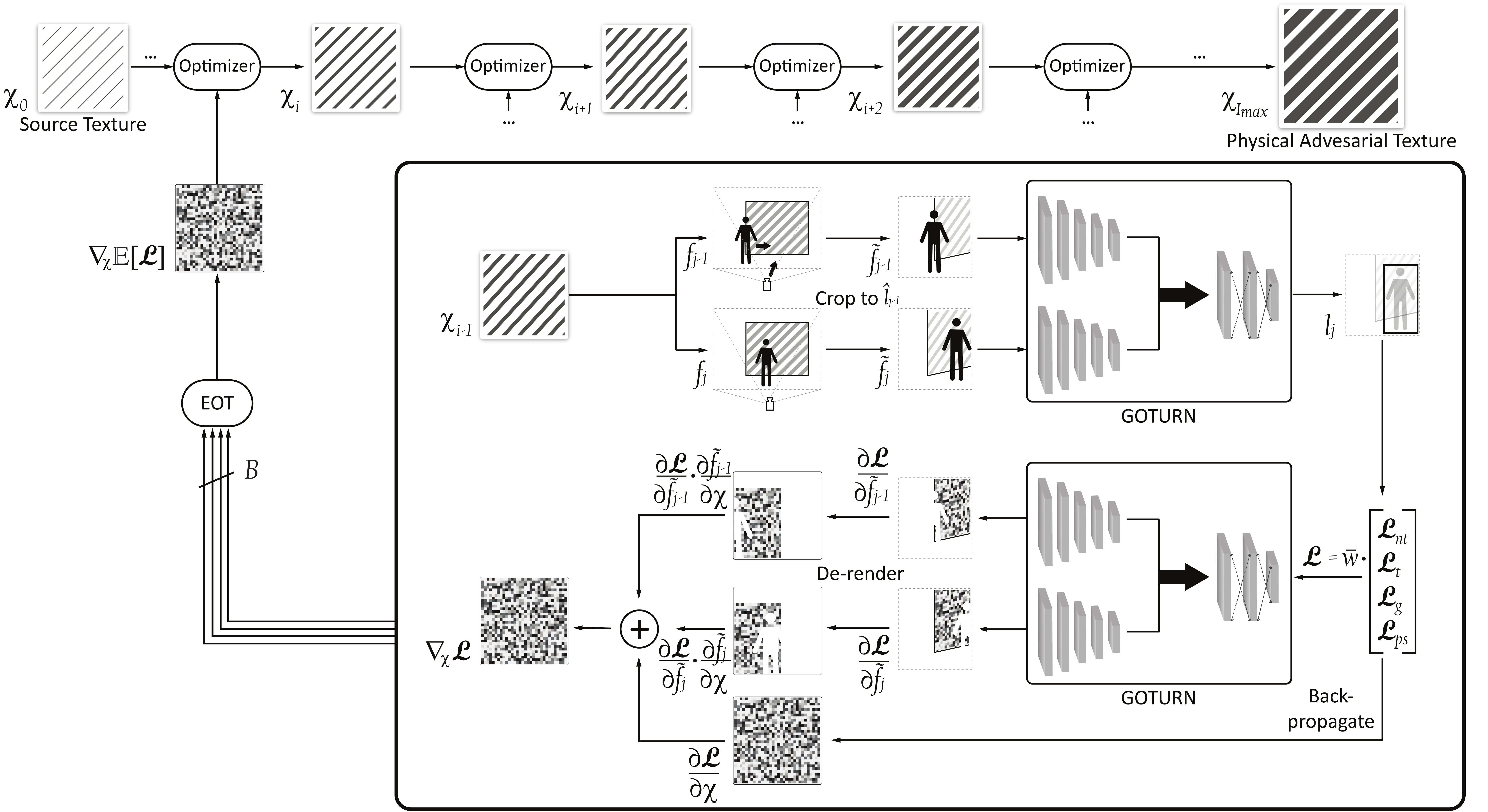}
\caption{The Physical Adversarial Texture (PAT) Attack creates adversaries to fool the GOTURN tracker, via minibatch gradient descent to optimize various losses, using randomized scenes following Expectation Over Transformation (EOT).}
\label{fig:block_diagram}
\vspace{-0.2cm}
\end{figure*}

In this work, we assume to have \emph{white-box} access to the GOTURN network's weights and thus the ability to back-propagate through it.
We focus on tracking people and humanoid robots in particular and assume that the tracker was trained on such types of targets.


As mentioned in Section~\ref{sec:intro}, several challenges arise when creating adversaries to fool temporal tracking models.
We address these by applying the Expectation Over Transformation (EOT) algorithm~\cite{pmlr-v80-athalye18b}, which minimizes the expected loss $\mathbb{E} \left[ \mathcal{L} \right]$ over a minibatch of $B$ scenes imaged under diverse conditions.
EOT marginalizes across the distributions of different transformation variables, such as the poses of the camera, tracked target, and poster, as well as the appearances of the target, environmental surroundings, and ambient lighting.
However, marginalizing over wide ranges of condition variables can be very computationally expensive.
Thus, Section~\ref{sec:eot_ablation} studies the effects on adversarial strength and attack speeds resulting from varying EOT variables.

An essential addition when generating a physical adversarial item, as opposed to a digital one, is the need to render the textured item into scenes as it evolves during the attack process.
Our attack creates PATs purely from scenes rendered using the Gazebo simulator~\cite{1389727}, yet Section~\ref{sec:sim2real} will show that these adversaries are also potent in the real world.


\subsection{Modeling rendering and lighting} \label{sec:lighting}

To optimize the loss with respect to the texture of a physical poster, we need to differentiate through the rendering process.
Rendering can be simplified into two steps: \emph{projecting} the texture onto the surface of a physical item and then onto the camera's frame, and \emph{shading} the color of each frame pixel depending on light sources and material types.

Similar to~\cite{liu2018beyond}, we sidestep shading complexities, such as spotlight gradients and specular surfaces, by assuming controlled imaging conditions: the PAT is displayed on a matte material and is lit by a far-away sun-like source, and the camera's exposure is adjusted not to cause pixel saturation.
Consequently, we employ a linear lighting model, where each pixel's RGB intensities in the camera frame is a scaled and shifted version of pixel values for the projected texture coordinate.
During our attack, we query the Gazebo simulation software to obtain exact gains for light intensity and material reflectance, while before each real-world test we fit parameters of this per-channel linear lighting model once, using a displayed color calibration target.
As for the projection component, we modified Gazebo's renderer to provide projected frame coordinates for each texture pixel (similar to~\cite{pmlr-v80-athalye18b}), as well as occlusion masks and bounding boxes of the target in the foreground.
We then use this texture-to-frame mapping to manually back-propagate through the projection process onto the texture space.

\subsection{PAT Attack}

Figure~\ref{fig:block_diagram} shows the overall procedure for generating a Physical Adversarial Texture.
Starting from a source texture $\chi_0$, we perform minibatch gradient descent on $\mathcal{L}$ to optimize pixel perturbations that adds onto the texture, for a total of $I_{max}$ iterations.
On each iteration $i$, we apply EOT to a minibatch of $B$ scenes, each with randomized settings for the poses of the camera, target, and poster, the identities of the target and background, and the hue-saturation-value settings of a single directional light source.

Each scene entails two frames $\{f_{j-1},f_{j}\}$, in which both the camera and tracked target may have moved between the previous and current frames.
Given the target's previous \emph{actual} location $\hat{l}_{j-1}$, we crop both frames around a correspondingly scaled region, then resize and process them through the GOTURN network, to predict the bounding-box location $l_j$ of the target in the current frame.
We then back-propagate from the combined loss objective $\mathcal{L}$ onto the texture space through all partial-derivative paths.
After repeating the above process for all $B$ scenes, we compute the expected texture gradient, and update the texture using the Fast Gradient Sign optimizer~\cite{43405}, scaled by the current iteration's step size $\alpha_i$:

\begin{equation}
\Delta \chi = - sign( \nabla_{\chi} \mathbb{E} \left[ \mathcal{L} \right] )
\end{equation}



\section{Experiments} \label{sec:experiments}

In this section, we present an empirical comparison of PAT attacks using non-targeted, targeted, guided, and hybrid losses.
We also assess which EOT conditioning variables are most useful for producing strong adversaries quickly.
Furthermore, we analyze PATs resulting from imitation attacks and their induced adversarial behaviors.
Finally, we showcase the transfer of PATs generated in simulation for fooling tracking system in a real-world setup.

\subsection{Setup} \label{sec:exp_setup}


All PAT attacks were carried out using simulated scenes rendered by Gazebo.
This conveniently provides an endless stream of independently-sampled scenes, with controlled poses and appearances for the target, textured poster, camera, background, and lighting.
We created multiple scenarios, including $3$ outdoor views of a $2.6\textrm{m} \times 2\textrm{m}$ poster in front of a building, forest, or playground, and an indoor coffee shop scene where a half-sized poster is hung on the wall.
We also varied tracked targets among models of $3$ different persons and $2$ humanoid robots.

\vspace{-0.1cm}
\subsubsection{Trained GOTURN models}

We trained several GOTURN networks on various combinations of synthetic and real-world labeled datasets for tracking people and humanoid robots.
The synthetic dataset contains over $1,400$ short tracking sequences with more than $300,000$ total frames, while the real-world dataset consists of $29$ videos with over $50,000$ frames of one of two persons, moving around an office garage and at a park.
We used the Adam optimizer~\cite{DBLP:journals/corr/KingmaB14} with an initial learning rate of $10^{-5}$ and a batch size of $32$.
Models trained on synthetic-only data (\texttt{sim}) lasted $300,000$ iterations with the learning rate halved every $30,000$ iterations, while those trained on combined datasets (\texttt{s+r}) or on the real-world dataset after bootstrapping from the synthetic-trained model (\texttt{s2r}) ran for $150,000$ iterations with the learning rate halved every $15,000$ iterations.
In addition to the architecture of~\cite{goturn2016} (\texttt{Lg}), we also trained smaller-capacity models with more aggressive striding instead of pooling layers and fewer units in the fully-connected layers (\texttt{Sm}).
While this section evaluates a subset of model instances, our Appendices present comprehensive results on other networks.

\subsubsection{Evaluation Metric} \label{sec:evaluation_metrics}

As discussed in Section~\ref{sec:adversarial_str}, we evaluate each PAT by generating sequences in which a tracked target moves from one side of the textured poster to the other.
Each sequence randomly draws from manually-chosen ranges for the target, camera, and poster poses, hue-saturation-value settings for the light source, target identities, and background scenes.
We run the GOTURN tracker on each sequence twice, differed by the display of either the PAT or an inert source texture on the poster.
Adversarial strength is then computed as the average $\mu IOUd$ metric over $20$ random sequence pairs.

Anecdotally, for average $\mu IOUd$ values around $0.2$, the tracker's predictions expanded and worsened as the target moved over the poster, yet GOTURN locked back onto the target as it moved away.
In contrast, values greater than $0.4$ reflected cases where GOTURN consistently lost track of the target during and at the end of the sequence, thus showing notably worse tracking compared to an inert poster.

\subsubsection{Baseline Attack Settings} \label{sec:baseline_config}

We carried out hyperparameter search to determine a set of attack parameters that produce strong adversaries (see Appendix~\ref{sec:appendix_baseline_config}).
Unless otherwise stated, each PAT attack ran on the regular-capacity synthetic-trained GOTURN model (\texttt{Lg,sim}), with: $I_{max}=1,000$ attack iterations, EOT minibatch with $B=20$ samples, FGS optimizer with step sizes of $\alpha_{i \le 500}=0.75$ and then $\alpha_{i > 500}=0.25$, and starting from a randomly-initialized source texture with $128 \times 128$ pixels.
All presented results are averaged over $10$ attack instances, with different initial random seeds.

\subsection{Efficacy of Adversarial Losses for Regression} \label{sec:exp_attack_variants}

\begin{figure}[h]
    \centering
    \begin{subfigure}{1.0\linewidth}
        \centering
        \includegraphics[width=\textwidth]{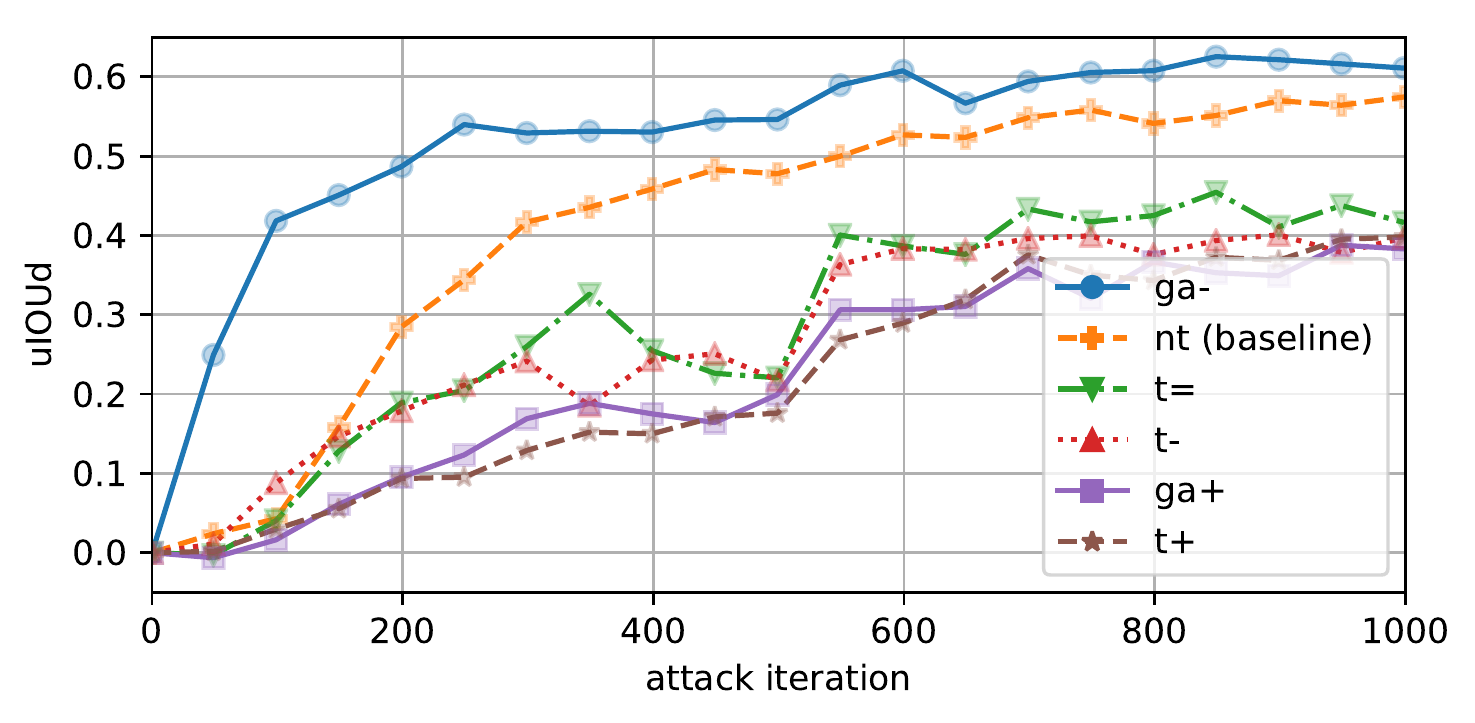}
        \caption{Different adversarial losses}
        \label{fig:variants_ind}
    \end{subfigure}
    \begin{subfigure}{1.0\linewidth}
        \centering
        \includegraphics[width=\textwidth]{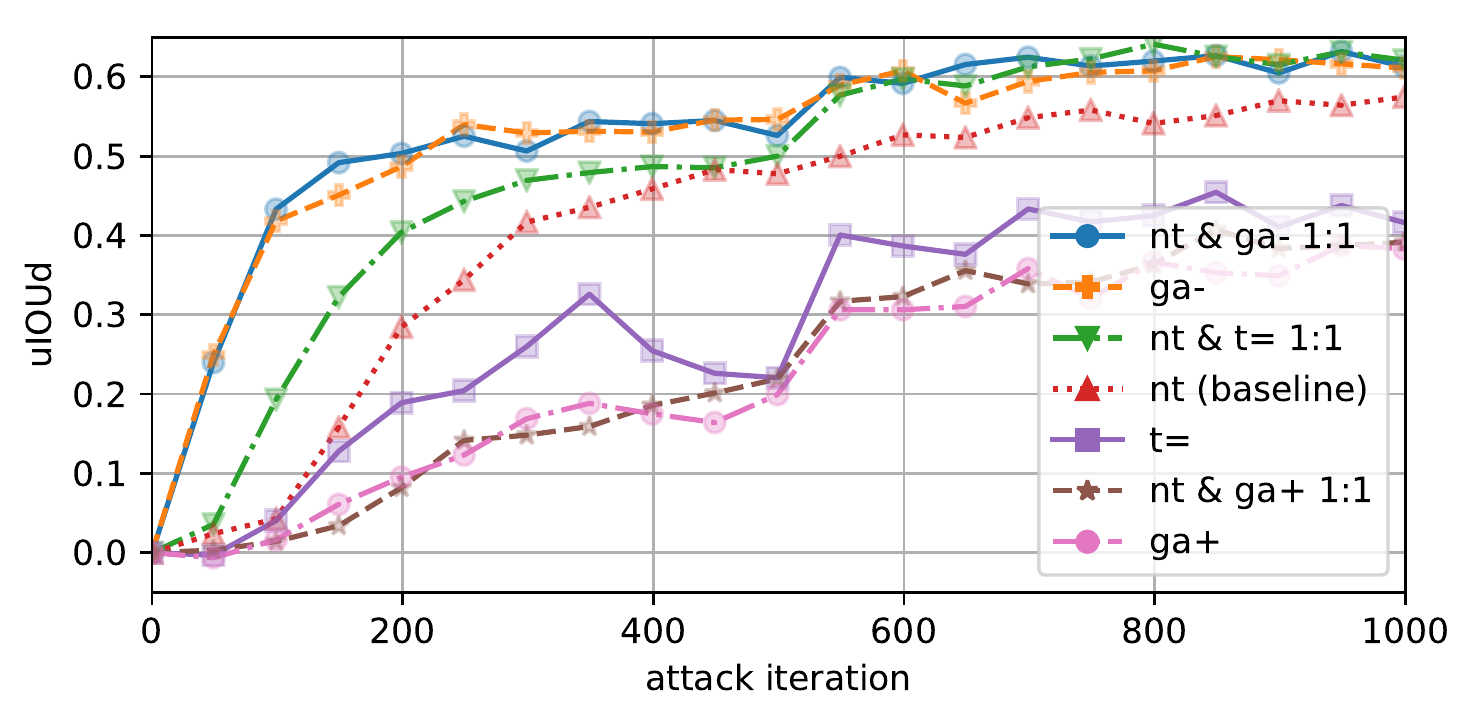}
        \caption{Individual vs hybrid adversarial losses}
        \label{fig:variants_hybrid}
    \end{subfigure}
    \caption{PAT attack strength for various adversarial losses.}
\label{fig:variants}
\end{figure}

Figure~\ref{fig:variants_ind} depicts the progression in adversarial strength throughout PAT attack runs for the different adversarial losses proposed in Section~\ref{sec:adversarial_loss}.
Comparing against the non-targeted baseline EOT attack ($\mathcal{L}_{nt}$)
, most targeted and guided losses resulted in slower convergence and worse final adversarial strength.
This is not surprising as these adversarial objectives apply stricter constraints on the desired adversarial behaviors and thus need to be optimized for longer.
As the sole exception, the guided loss encouraging smaller-area predictions ($\mathcal{L}_{ga-}$)
attained the fastest convergence and best adversarial strength overall.
This suggests that \emph{well-engineered adversarial objectives, especially loosely-guided ones, benefit by speeding up and improving the attack process} on regression tasks.

In Figure~\ref{fig:variants_hybrid}, we see that combining $\mathcal{L}_{nt}$ with most targeted or guided losses did not significantly change performance.
While not shown, we saw similar results when using 1:1000 weight ratios.
However, the 1:1 combination of $\mathcal{L}_{nt} \ \& \ \mathcal{L}_{t=}$ attained better overall performance than both $\mathcal{L}_{nt}$ and $\mathcal{L}_{t=}$.
This suggests that \emph{sometimes adding a non-targeted loss to a targeted or guided one helps}, possibly due to the widening of conditions for adversarial behaviors.

\begin{figure}[h!]
\centering
\includegraphics[width=1.0\linewidth]{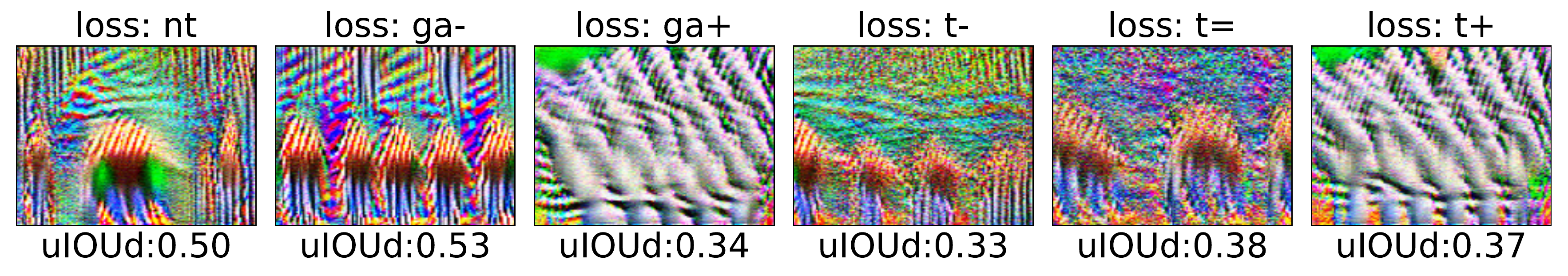}
\caption{PATs generated using different adversarial losses.}
\label{fig:attack_variants_results}
\end{figure}

As seen in Figure~\ref{fig:attack_variants_results}, various patterns emerge in PATs generated by different losses.
We note that dark \enquote{striped patches} always appeared in PATs generated from certain losses, and these patches caused GOTURN to lock on and break away from the tracked target.
On the other hand, \enquote{striped patches} did not show up for PATs created using $\mathcal{L}_{ga+}$ 
or $\mathcal{L}_{t+}$ 
, which showed uniform patterns.
This is expected as these losses encourage the tracker's predictions to grow in size, rather than fixating onto a specific location.

\subsection{Ablation of EOT Conditioning Variables} \label{sec:eot_ablation}

Here, we assess which variables for controlling the random sampling of scenes had strong effects, and which ones could be set to fixed values without impact, thus reducing scene randomization and speeding up EOT-based attacks.

\begin{figure}[ht]
    \centering
    \begin{subfigure}{1.0\linewidth}
        \centering
        \includegraphics[width=\textwidth]{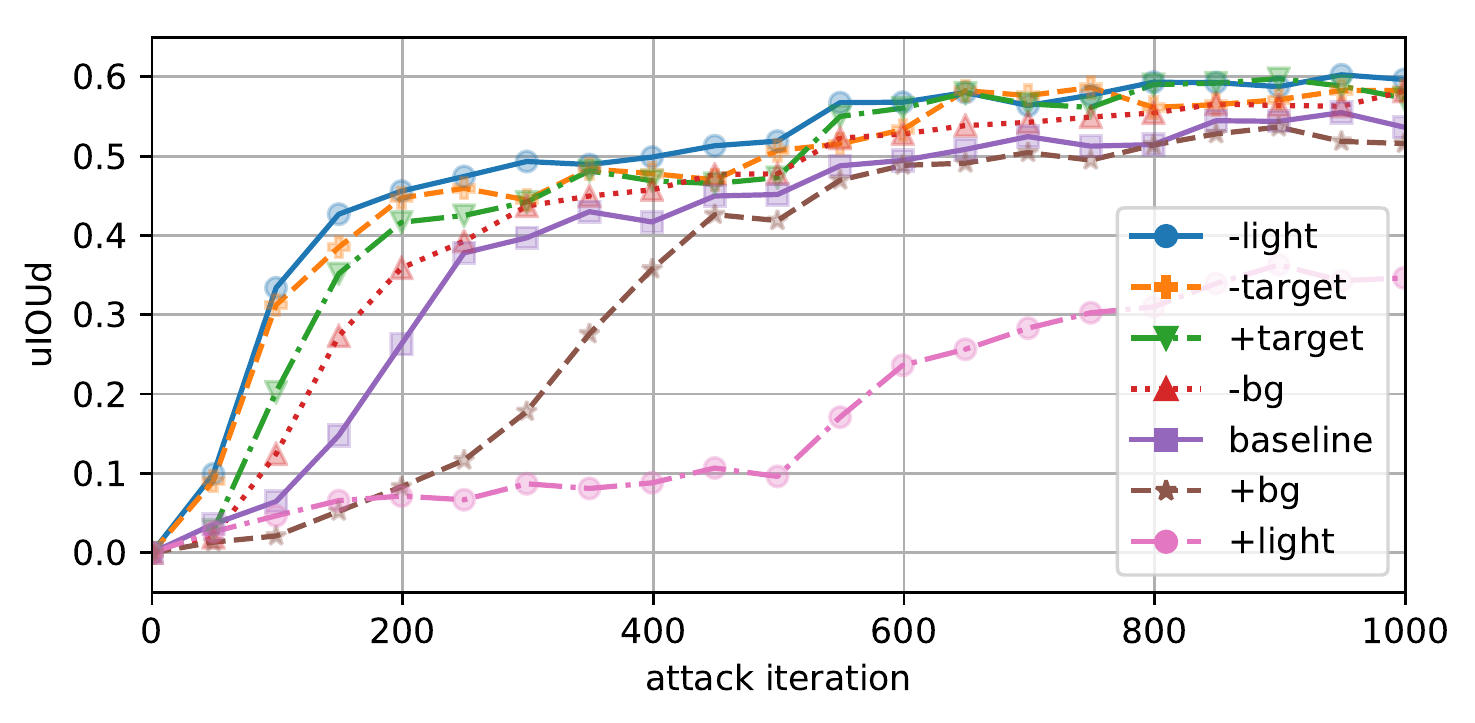}
        \caption{Variables controlling randomized appearances}
        \label{fig:eot_ablation_appearance}
    \end{subfigure}
    \begin{subfigure}{1.0\linewidth}
        \centering
        \includegraphics[width=\textwidth]{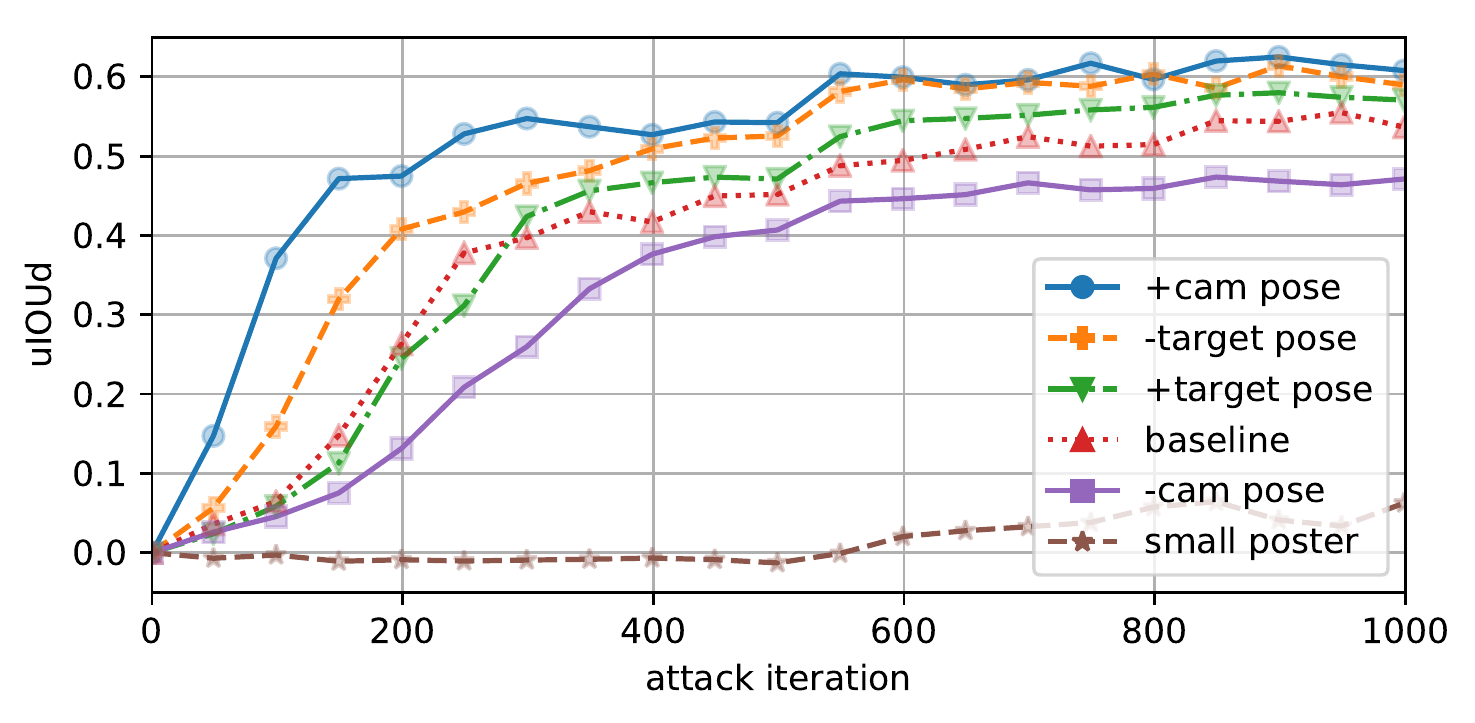}
        \caption{Variables controlling randomized poses}
        \label{fig:eot_ablation_position}
    \end{subfigure}
\caption{PAT attack strength for various EOT variables.}
\label{fig:eot_ablation}
\vspace{-0.2cm}
\end{figure}

As seen in Figure~\ref{fig:eot_ablation_appearance}, reducing variety in appearances of the background (\texttt{-bg}), target (\texttt{-target}), and light variations (\texttt{-light}), did not substantially affect adversarial strength when other parameter ranges were held constant.
Also, increasing diversity in \texttt{+target} and \texttt{+bg} did not result in different end-performance.
This suggests that \emph{diversity in target and background appearances do not strongly affect EOT-based attacks}.
On the other hand, \texttt{+light} converged much slower than other settings.
Thus, we conclude that \emph{if randomized lighting is needed to generalize the robustness of PATs during deployment, then more attack iterations are needed to ensure convergence}.

For pose-related variables in Figure~\ref{fig:eot_ablation_position}, halving the poster size (\texttt{small poster}) caused the PAT attack to fail.
Changing the ranges of camera poses (\texttt{+cam pose}, \texttt{-cam pose}) resulted in notable performance differences, therefore we note that \emph{more iterations are needed to generate effective PATs under wider viewpoint ranges}.
Perhaps surprisingly, for \texttt{-target pose}, \emph{locking the target's pose to the center of the poster resulted in faster and stronger convergence}.
This is likely because regions around the static target obtained consistent perturbations across all scenes, and so developed adversarial patterns faster.

\subsection{Imitation Attacks} \label{sec:imitation}

As discussed in Section~\ref{sec:adversarial_loss}, we can add a perceptual similarity loss term to make the PAT imitate a meaningful source image.
A larger perceptual similarity weight $w_{ps}$ perturbs the source less, but at the cost of slower convergence and weaker or ineffective adversarial strength.
Results below reflect a manually-tuned setting of $w_{ps}=0.6$.



\begin{figure}[h]
\centering
\includegraphics[width=1.0\linewidth]{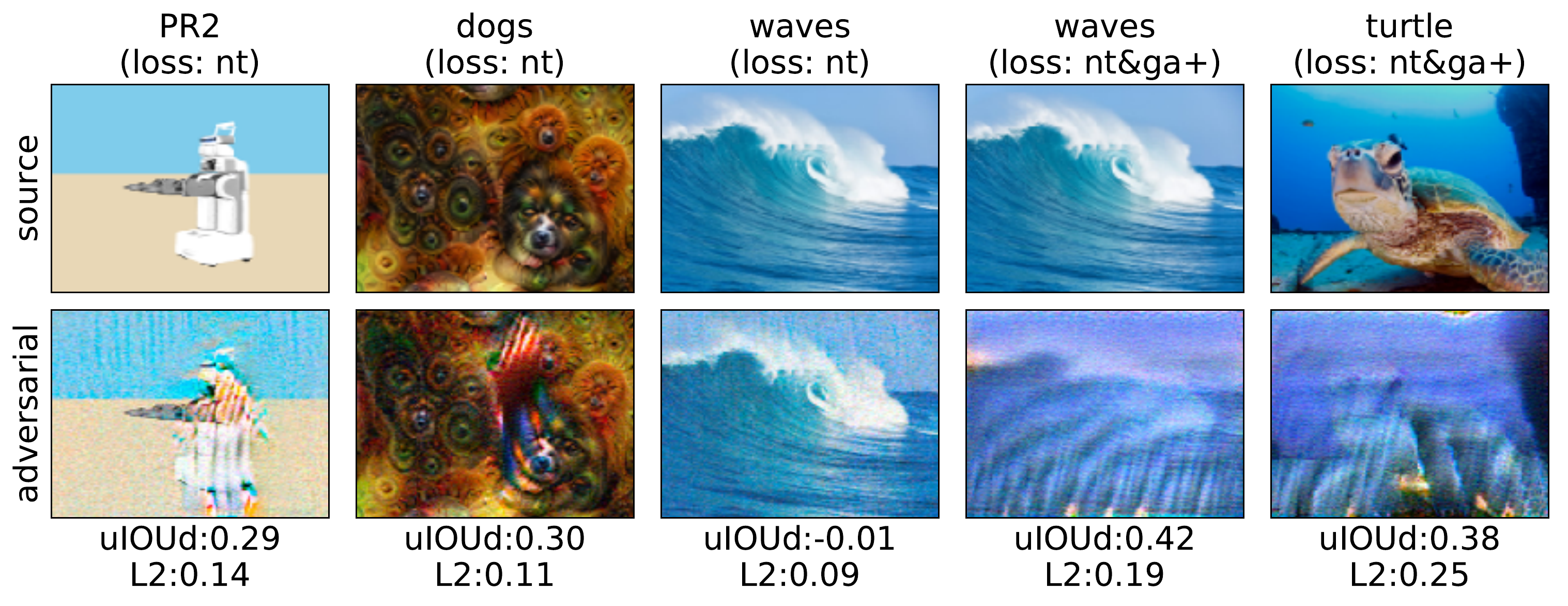}
\caption{Adversarial imitations under various losses.}
\label{fig:imitation_samples}
\vspace{-0.15cm}
\end{figure}

Figure~\ref{fig:imitation_samples} shows that some source images, coupled with the right adversarial loss, led to stronger imitations than others.
For instance, the \texttt{waves} source was optimized into a potent PAT using $\mathcal{L}_{nt} \ \& \ \mathcal{L}_{ga+}$, yet using $\mathcal{L}_{nt}$ alone failed to produce an adversarial texture.
However, we found that for a given threshold on $L_2$ distance, guided losses generally converged faster to reach potent behaviors, yet suffered from weakened adversarial strength compared to $\mathcal{L}_{nt}$ over prolonged attack iterations (see Appendix~\ref{sec:appendix_imitation} for quantitative details).
Also, under larger $w_{ps}$ constraints, we saw that adversarial perturbations appeared only in selective parts of the texture.
Notably, the \enquote{striped patches} seen in non-imitated PATs (Figure~\ref{fig:attack_variants_results}) also emerged near the \texttt{dogs}' face and over the \texttt{PR2} robot, when optimized using $\mathcal{L}_{nt}$.
We thus conclude that the PAT attack produces \emph{critical adversarial patterns} such as these patches first, and then perturbs other regions into \emph{supporting adversarial patterns}. 

\begin{figure}[hb]
    \centering
    \begin{subfigure}{0.23\textwidth}
        \centering
        \includegraphics[width=\textwidth]{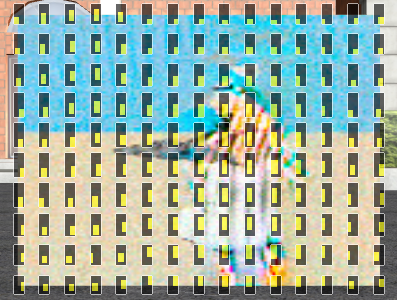}
        \caption{\texttt{Lg,sim} tracker; $\mathcal{L}_{nt}$ loss}
        \label{fig:imitation_pr2}
    \end{subfigure}
    \begin{subfigure}{0.23\textwidth}
        \centering
        \includegraphics[width=\textwidth]{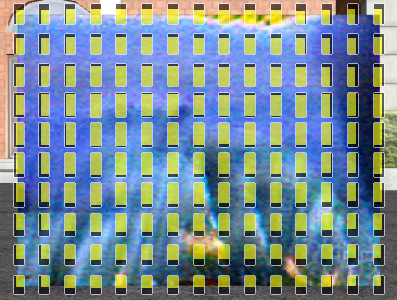}
        \caption{\texttt{Lg,s+r} tracker; $\mathcal{L}_{ga+}$ loss}
        \label{fig:imitation_turtle}
    \end{subfigure}
    \caption{Adversarial behaviors emerging from PATs.}
    \label{fig:imitation}
\end{figure}

Further substantiating this claim, Figure~\ref{fig:imitation} visualizes predicted bounding-boxes within search areas located at different sub-regions of PATs.
We see from Figure~\ref{fig:imitation_pr2} that predictions around the adversarial \enquote{striped patch} made GOTURN track towards it.
This suggests that such \emph{critical adversarial patterns induce potent lock-on behaviors that break tracking, regardless of where the actual target is positioned}.
On the other hand, shown in Figure~\ref{fig:imitation_turtle}, the \enquote{regular wavy} pattern optimized using $\mathcal{L}_{ga+}$ resulted in the intended adversarial behavior of larger-sized predictions, regardless of the search area's location.

\subsection{Demonstration of Sim-to-real Transfer} \label{sec:sim2real}





To assess the real-world effectiveness of PATs generated purely using simulated scenes, we displayed them on a $50''$ TV within an indoor environment with static lighting.
We carried out two sets of person-following experiments using the camera on a Parrot Bebop 2 drone: \emph{tracking} sessions with a stationary drone, and \emph{servoing} runs where the tracked predictions were used to control the robot to follow the target through space (see Section~\ref{sec:goturn} for details).

In both experiments, we tasked the \texttt{s+r} GOTURN instance to follow people that were not seen in the tracker's training dataset.
While we tested under different light intensities, for each static setting, we first fit a linear per-channel lighting model to a color calibration target, and then adjusted camera frames accordingly, as explained in Section~\ref{sec:lighting}.
We carried out this \emph{optional} step to showcase adversarial performance in best-case conditions, and note that none of the simulated evaluations corrected for per-scenario lighting.
Also, this correction compensates for fabrication errors that may arise when displaying the PAT on a TV or printed as a static poster, and further serves as an alternative to adding a Non-Printability Score to the attack loss~\cite{Sharif:2016:ACR:2976749.2978392}.

\begin{figure}[b]
    \centering
    \includegraphics[width=1.0\linewidth]{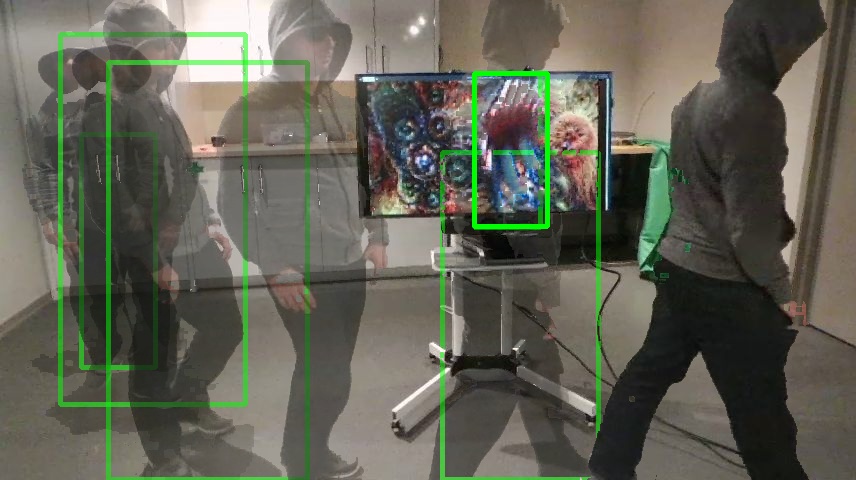}
    \caption{An imitated PAT, created in simulation, can fool a person-tracker in the real world.
    }
    \label{fig:real_pat_sample}
\end{figure}

During our experiments, we observed $57/80$ stationary runs and $6/18$ servoing runs to have strong lock-on adversarial behaviors.
For succinctness, we focus on qualitative analyses below; please refer to Appendix~\ref{sec:appendix_sim2real} for more extensive quantitative results and visual samples.

For stationary tracking runs, only adversaries containing \enquote{striped patches} consistently made GOTURN break away from the person.
Other PATs optimized by, e.g., $\mathcal{L}_{ga+}$, caused the tracker to make worse predictions as the target moved in front of the poster, yet it ultimately locked back onto the person.
While these results were partially due to our limited-size digital poster, a more general cause is likely because such losses induced weak adversarial behaviors: by encouraging growing predictions, GOTURN could still see and thus track the person within an enlarged search area.

Returning to the best-performing PATs containing \enquote{striped patches}, the tracker strongly preferred to lock onto these rather than the person.
Moreover, even though the person could regain GOTURN's focus by completely blocking the patch, as soon as he or she moved away, the tracker locked back onto the patch, as seen in Figure~\ref{fig:real_pat_sample}.
Furthermore, these physical adversaries were robust to various viewing distances and angles, and even for settings outside the ranges used to randomize scenes during the PAT attack.

Our servoing tests showed that it was generally harder to make GOTURN completely break away from the target.
Since the drone was moving to follow the target, even though the tracker's predictions were momentarily disturbed or locked onto the PAT, often the robot's momentum caused GOTURN to return its focus onto the person.
We attribute the worsened PAT performance to motion blurring, light gradients, and specular reflections that were present due to the moving camera, all of which were assumed away by our PAT attack.
Nevertheless, we believe that these advanced scene characteristics can be marginalized by the EOT algorithm, using a higher-fidelity rendering engine than our implementation.

Finally, we speculate that synthetically-generated adversarial patterns like the \enquote{striped patches} may look like simulated people or robot targets in GOTURN's view.
If so, then our real-world transfer experiments may have been aided by GOTURN's inability to tell apart synthetic targets from real people.
This caveat may be overcome by carrying out PAT attack using scenes synthesized with textured 3-D reconstructions or photograph appearances of the intended target.



\section{Conclusion} \label{sec:conclusion}

We presented a system to generate Physical Adversarial Textures (PAT) for fooling object trackers.
These \enquote{PATterns} induced diverse adversarial behaviors, emerging from a common optimization framework with the end-goal of making the tracker break away from its intended target.
We compared different adversarial objectives and showed that a new family of guided losses, when well-engineered, resulted in stellar adversarial strength and convergence speed.
We also showed that a naive application of EOT by randomizing \emph{all} aspects of scenes was not necessary.
Finally, we showcased synthetically-generated PATs that can fool real-world trackers.


We hope to raise awareness that inconspicuously-colored items can mislead modern vision-based systems \emph{by merely being present in their vicinity}.
Despite recent 
advances, we argue that purely vision-based tracking systems are not robust to physical adversaries, and thus recommend commercial tracking and servoing systems to integrate auxiliary signals (e.g., GPS and IMU) for redundancy and safety.

Since a vital goal of this work is to show the existence of inconspicuous patterns that fool trackers, we made the simplifying assumption of white-box access. 
More practically, it might be possible to augment the PAT attack using diverse techniques~\cite{Papernot:2017:PBA:3052973.3053009,chen2017zoo,8601309} to fool black-box victim models.
Another improvement could be to directly optimize non-differentiable
metrics such as $\mu IOUd$ by, e.g., following the Houdini method~\cite{NIPS2017_7273}.
Finally, although the textures shown in this work may appear inconspicuous prior to our demonstrations, they are nevertheless clearly visible and thus can be detected and protected against.
As the research community aims to defend against 
physical 
adversaries, we should continue to be on the lookout for potent PATs that more closely imitate natural items in the physical world.




\section*{Acknowledgements}

We want to thank Dmitri Carpov, Matt Craddock, and Ousmane Dia for helping on the codebase implementation, Nicolas Chapados and Pedro Pinheiro for valuable feedback on our manuscript, and Minh Dao for helping with visual illustrations. We would also like to thank Philippe Beaudoin, Jean-Fran\c{c}ois Marcil, and Sharlene McKinnon for participating in our real-world 
experiments.

{\small
\bibliographystyle{ieee_fullname}
\bibliography{references}
}

\begin{appendices}

\section{Simulated Scenarios}

\begin{figure}[h]
    \centering
    \begin{subfigure}{0.49\linewidth}
        \centering
        \includegraphics[width=\textwidth]{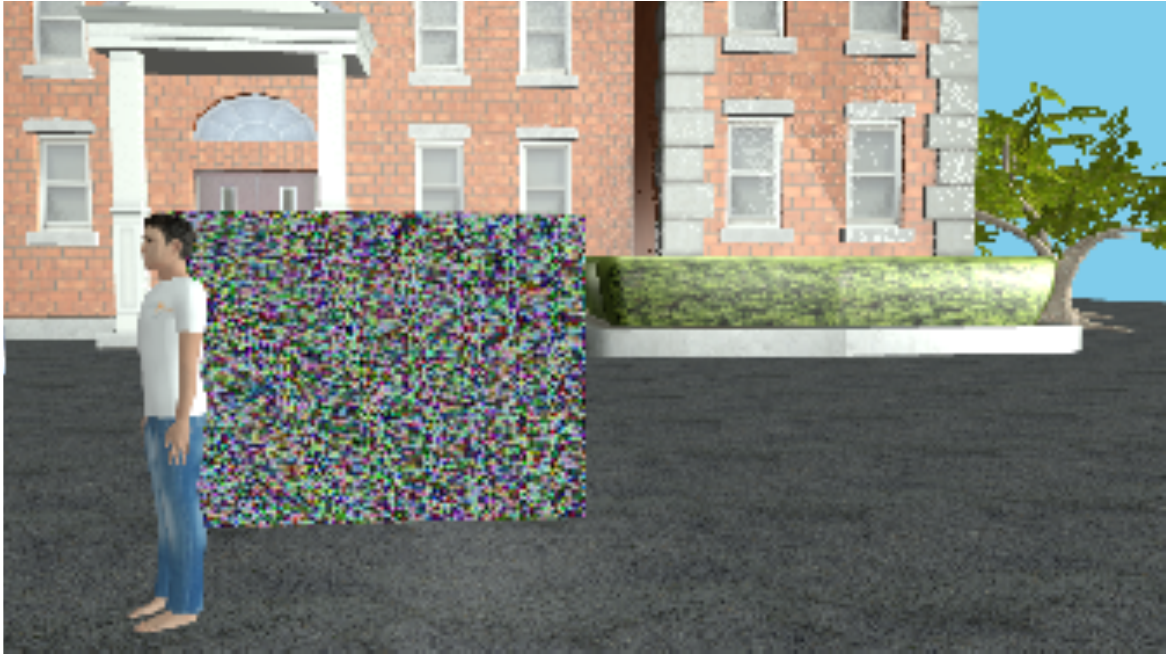}
        \caption{\texttt{t-shirt person} in \texttt{school}}
        \label{fig:sam le_1}
    \end{subfigure}
    \begin{subfigure}{0.49\linewidth}
        \centering
        \includegraphics[width=\textwidth]{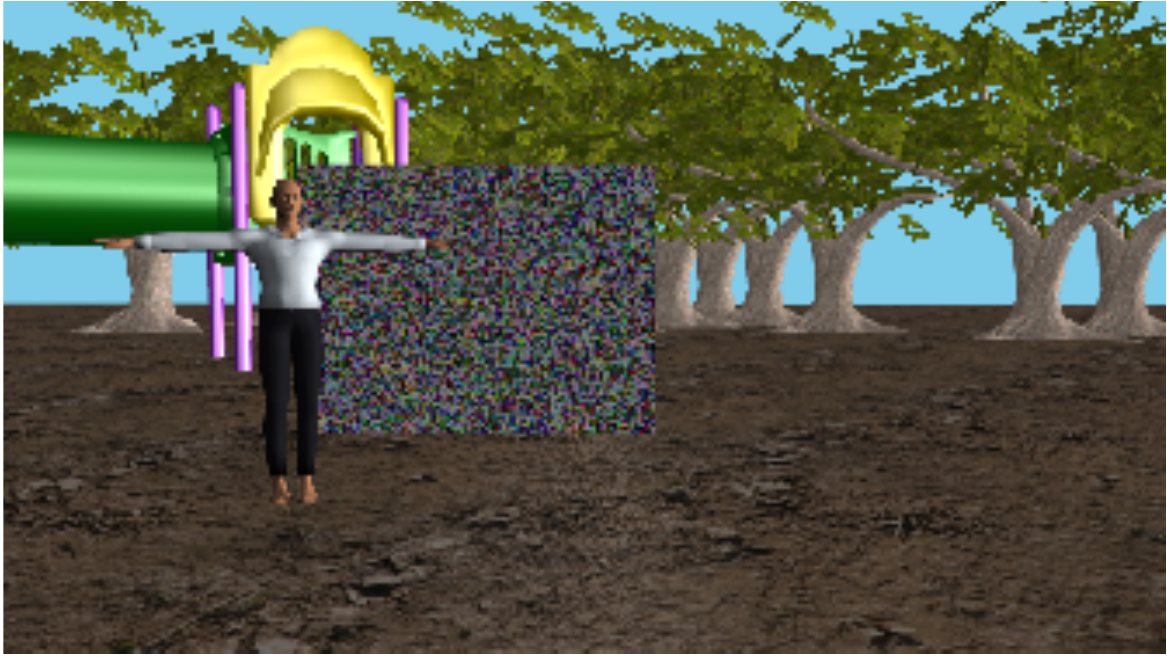}
        \caption{\texttt{white person} in \texttt{playground}}
        \label{fig:sample_2}
    \end{subfigure}
    \begin{subfigure}{0.49\linewidth}
        \centering
        \includegraphics[width=\textwidth]{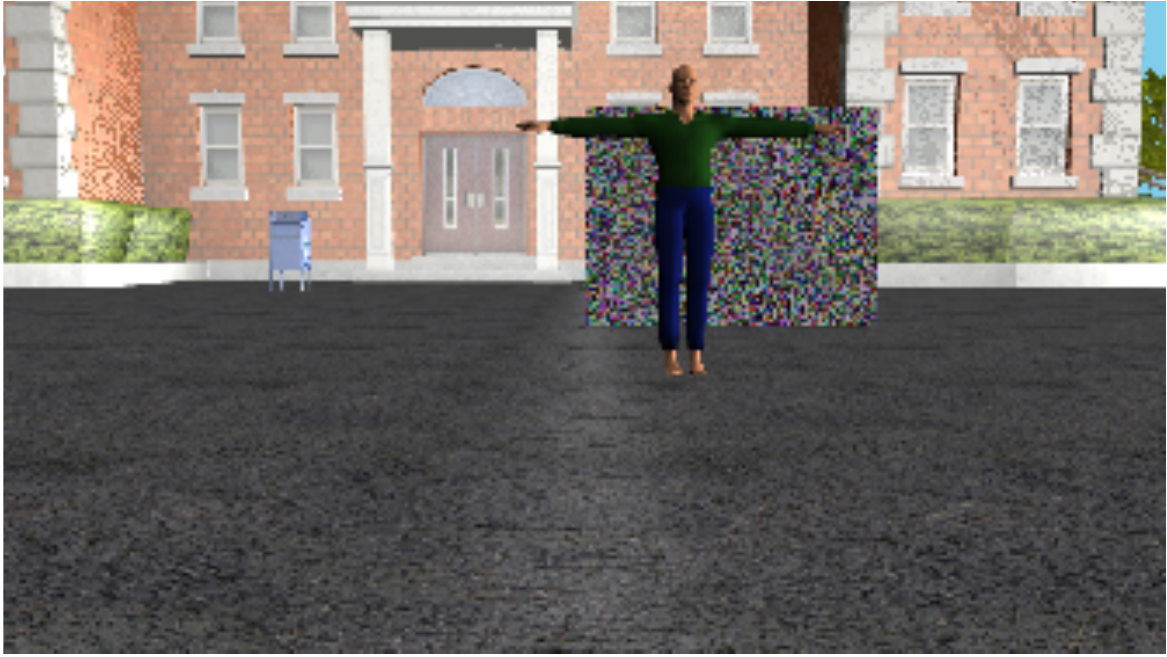}
        \caption{\texttt{green person} in \texttt{school}}
        \label{fig:sample_3}
    \end{subfigure}
    \begin{subfigure}{0.49\linewidth}
        \centering
        \includegraphics[width=\textwidth]{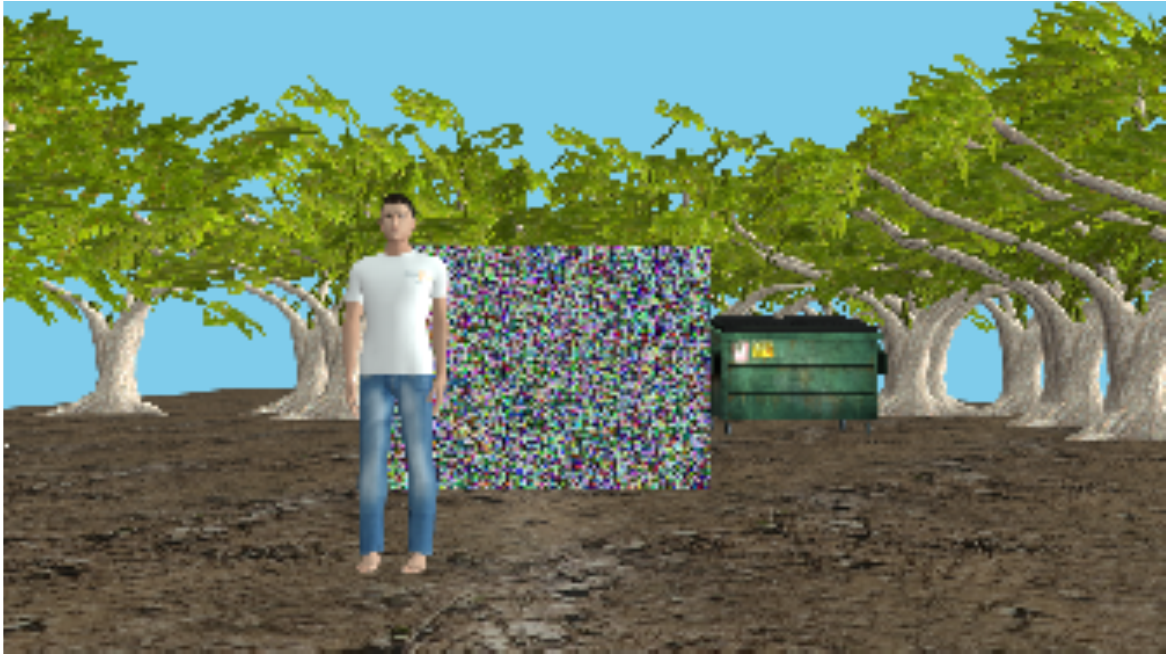}
        \caption{\texttt{t-shirt person} in \texttt{forest}}
        \label{fig:sample_4}
    \end{subfigure}
    \begin{subfigure}{0.49\linewidth}
        \centering
        \includegraphics[width=\textwidth]{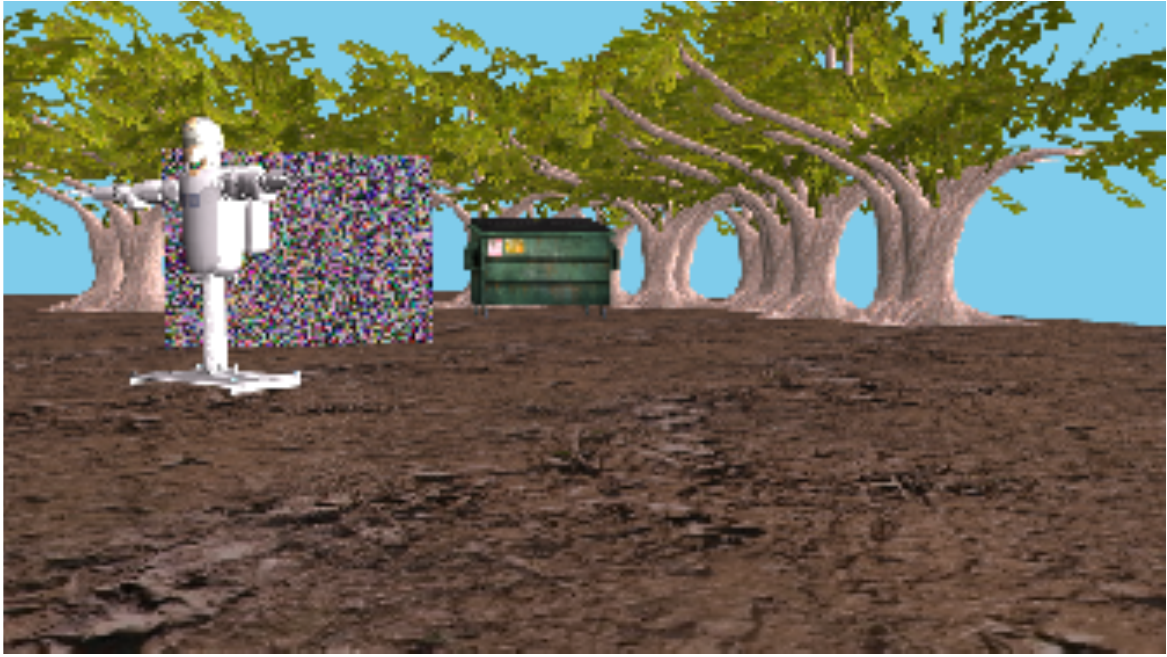}
        \caption{\texttt{robonaut} in \texttt{forest}}
        \label{fig:sample_5}
    \end{subfigure}
    \begin{subfigure}{0.49\linewidth}
        \centering
        \includegraphics[width=\textwidth]{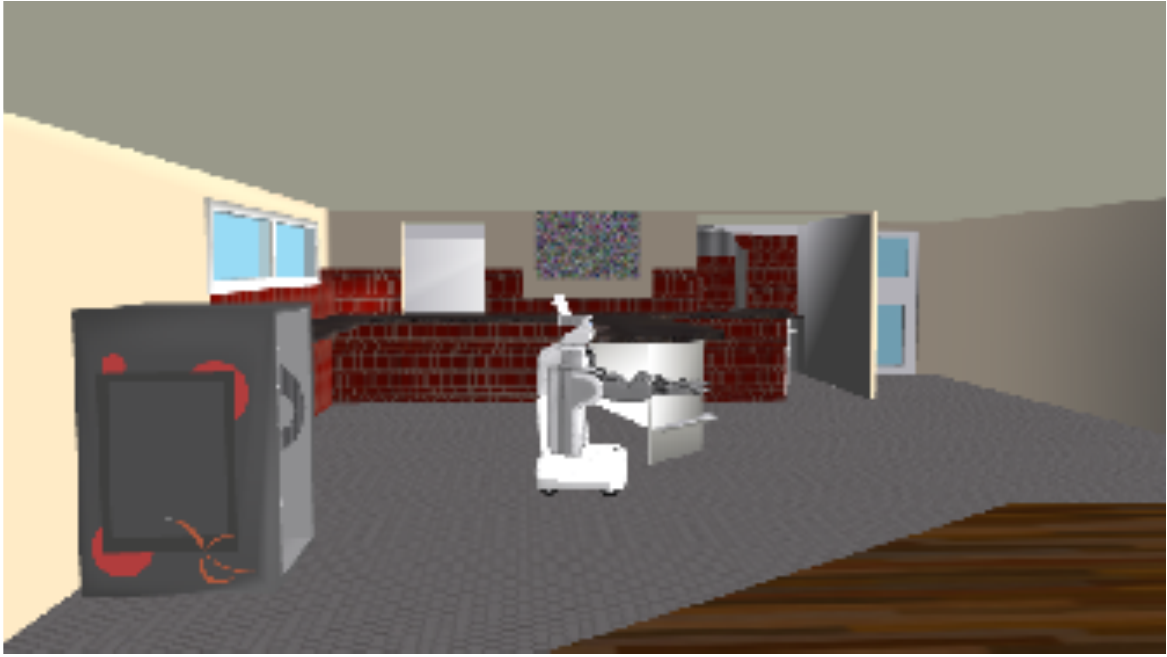}
        \caption{\texttt{PR2} in \texttt{cafe}}
        \label{fig:sample_6}
    \end{subfigure}
\caption{Samples of simulated scenarios.}
\label{fig:sample_scenarios}
\end{figure}

Figure~\ref{fig:sample_scenarios} depicts samples of the $5$ target (human or humanoid robot models) and $4$ scenarios (outdoor and indoor scenes) that we created within the Gazebo simulation software~\cite{1389727}.
These are used both for generating and evaluating Physical Adversarial Textures (PAT).

\section{PAT Attack: Random Scene Configuration} \label{sec:appendix_default_transformation}

The Expectation Over Transformation (EOT) algorithm~\cite{pmlr-v80-athalye18b} randomizes various parameters and aspects of scenes, such as camera placement and target appearance. By optimizing on these diverse and randomized scenes, we can ensure that the generated PAT would likely be universally adversarial.
Table~\ref{table:default_transformation} presents \emph{default} ranges used for continuous transformation variables used in our PAT Attack process, while Table~\ref{table:default_background_target} enumerates selections for discrete transformation variables.
This default configuration is used in Sections~\ref{sec:exp_attack_variants},~\ref{sec:imitation}, and~\ref{sec:sim2real}.

\begin{table}[ht]
\centering
\caption{Continous EOT variable ranges for PAT attack.}
\begin{tabular*}{0.38 \textwidth}{lrr}
\hline
Transformation & Min & Max \\
\hline
Initial camera x (m) & -1.5 & 1.5 \\
Initial camera y (m) & -11.0 & -6.0 \\
Initial camera z (m) & 0.6 & 1.8 \\
Initial camera roll (\degree) & 0.0 & 0.0 \\
Initial camera pitch (\degree) & -5.0 & 5.0 \\
Initial camera yaw (\degree) & -15.0 & 15.0 \\
\hline
Camera $\Delta$x (m) & -0.1 & 0.1 \\
Camera $\Delta$y (m) & -0.5 & 0.5 \\
Camera $\Delta$z (m) & -0.1 & 0.1 \\
Camera $\Delta$roll (\degree) & 0.0 & 0.0 \\
Camera $\Delta$pitch (\degree) & -3.0 & 3.0 \\
Camera $\Delta$yaw (\degree) & -3.0 & 3.0 \\
\hline
Initial target x (m) & -1.4 & 1.4 \\
Initial target y (m) & -5.0 & -0.7 \\
Initial target z (m) & 0.0 & 0.0 \\
Initial target roll (\degree) & 0.0 & 0.0 \\
Initial target pitch (\degree) & 0.0 & 0.0 \\
Initial target yaw (\degree) & 0.0 & 180.0 \\
\hline
Target $\Delta$x (m) & -0.1 & 0.1 \\
Target $\Delta$y (m) & -0.1 & 0.1 \\
Target $\Delta$z (m) & 0.0 & 0.0 \\
Target $\Delta$roll (\degree) & 0.0 & 0.0 \\
Target $\Delta$pitch (\degree) & 0.0 & 0.0 \\
Target $\Delta$yaw (\degree) & -10.0 & 10.0 \\
\hline
Lighting diffuse hue & 0.0 & 360.0 \\
Lighting diffuse saturation & 0.0 & 0.2 \\
Lighting diffuse value & 0.1 & 0.7 \\
\hline
\end{tabular*}
\label{table:default_transformation}
\end{table}

\begin{table}[ht]
\centering
\caption{Discrete EOT variable selections for PAT attack.}
\begin{tabular*}{0.3 \textwidth}{rl}
\hline
Backgrounds & Targets \\
\hline
\texttt{school} & \texttt{green person} \\
\texttt{forest} & \texttt{PR2} \\
\hline
\end{tabular*}
\label{table:default_background_target}
\end{table}

\begin{figure*}[ht]
    \centering
    \begin{subfigure}{1.0\linewidth}
        \centering
        \includegraphics[width=\textwidth]{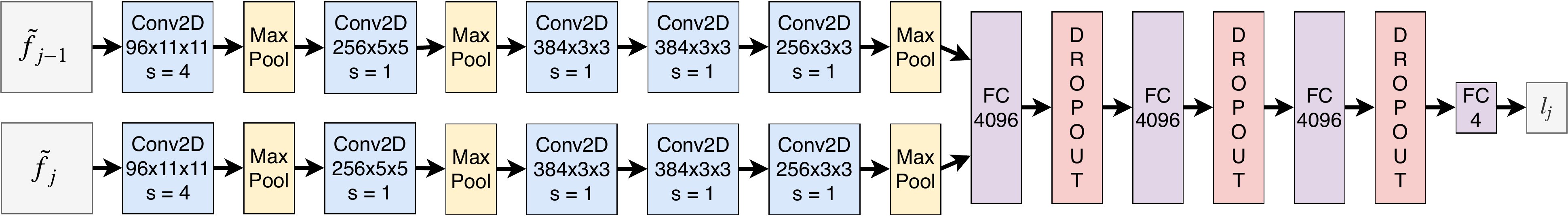}
        \caption{Regular-capacity model~\cite{goturn2016} (\texttt{Lg})}
        \label{fig:goturn_large}
    \end{subfigure}
    \begin{subfigure}{0.6\linewidth}
        \centering
        \includegraphics[width=\textwidth]{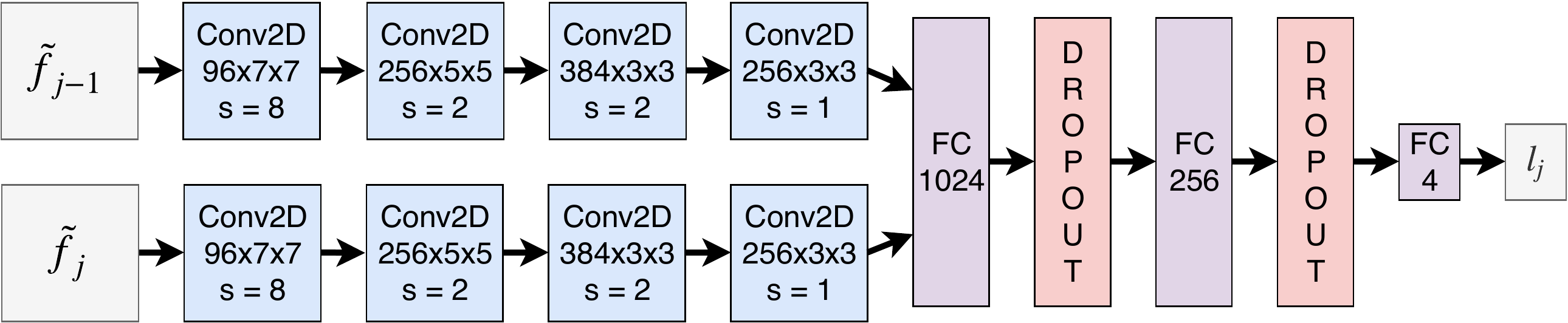}
        \caption{Reduced-capacity model (\texttt{Sm})}
        \label{fig:goturn_small}
    \end{subfigure}
\caption{Neural architectures for the GOTURN object tracker instances.}
\label{fig:goturn_arch}
\end{figure*}

\section{Trained GOTURN models}

Figure~\ref{fig:goturn_arch} illustrates the two GOTURN neural object tracking architectures used in our experiments.

\section{Baseline PAT Attack Settings} \label{sec:appendix_baseline_config}

The parameters used in the baseline PAT attack settings (see Section~\ref{sec:baseline_config}) were determined using hyperparameters search, and from conducting sensitivity analyses on EOT minibatch size and iteration, as well as texture attributes experiments.

\subsection{EOT Minibatch Size and Iteration} \label{sec:appendix_sensitiviy_eot}

Similar to how training a neural network using Stochastic Gradient Descent (SGD) is sensitive to hyperparameter settings, we analyzed the sensitivity of our proposed PAT attack method to its hyperparameters.
We suspect that attacks using smaller EOT minibatch sizes $B$ would require more iterations $I$ to converge assuming a fixed perturbation step size $\alpha$, while attacks using large minibatch sizes $B$ would require an impractical amount of computing time per iteration.
Thus, it is practically beneficial to balance the combination of the perturbation step size $\alpha$ and the minibatch size $B$, given a fixed number of attack iterations $I$.

\begin{figure}[h]
    \centering
    \begin{subfigure}{1.0\linewidth}
        \centering
        \includegraphics[width=\textwidth]{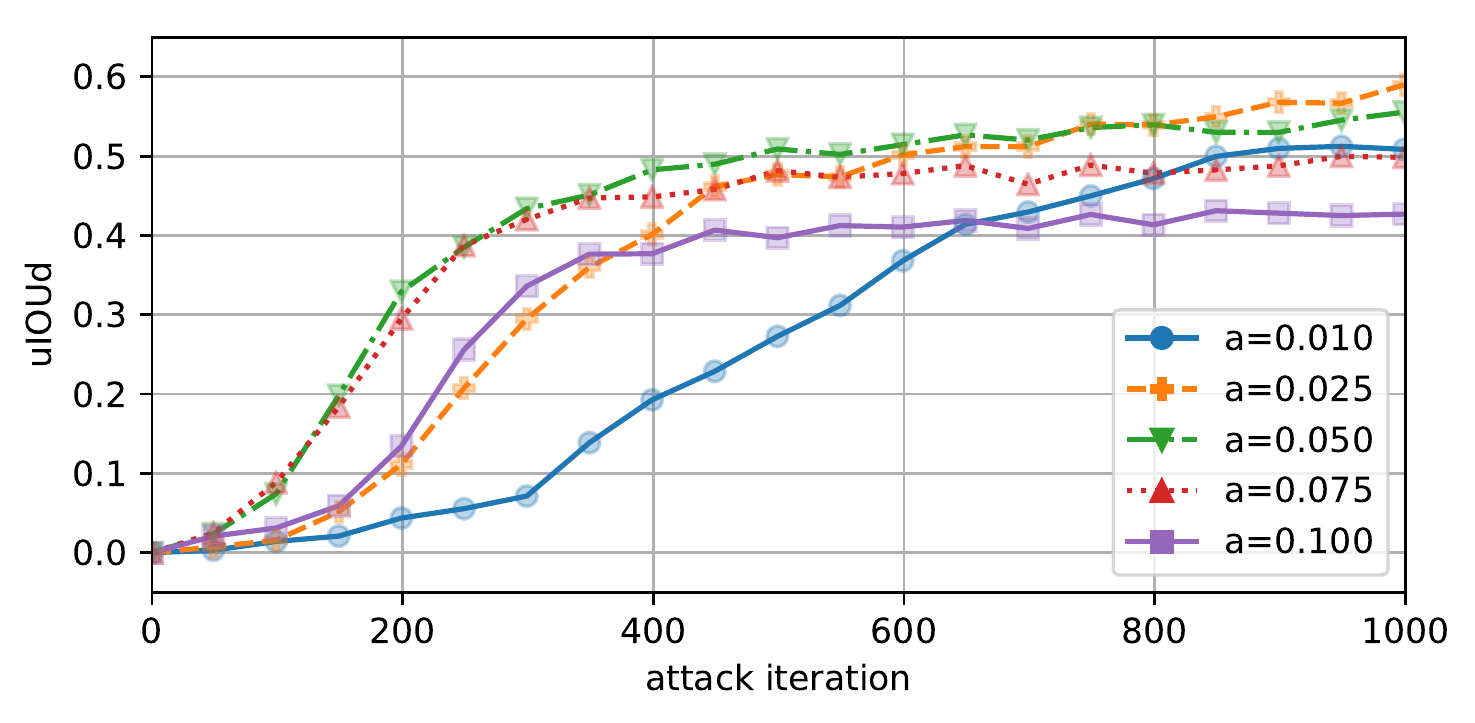}
        \caption{Perturbation step size $\alpha$}
        \label{fig:sensitivity_atkiters}
    \end{subfigure}
    \begin{subfigure}{1.0\linewidth}
        \centering
        \includegraphics[width=\textwidth]{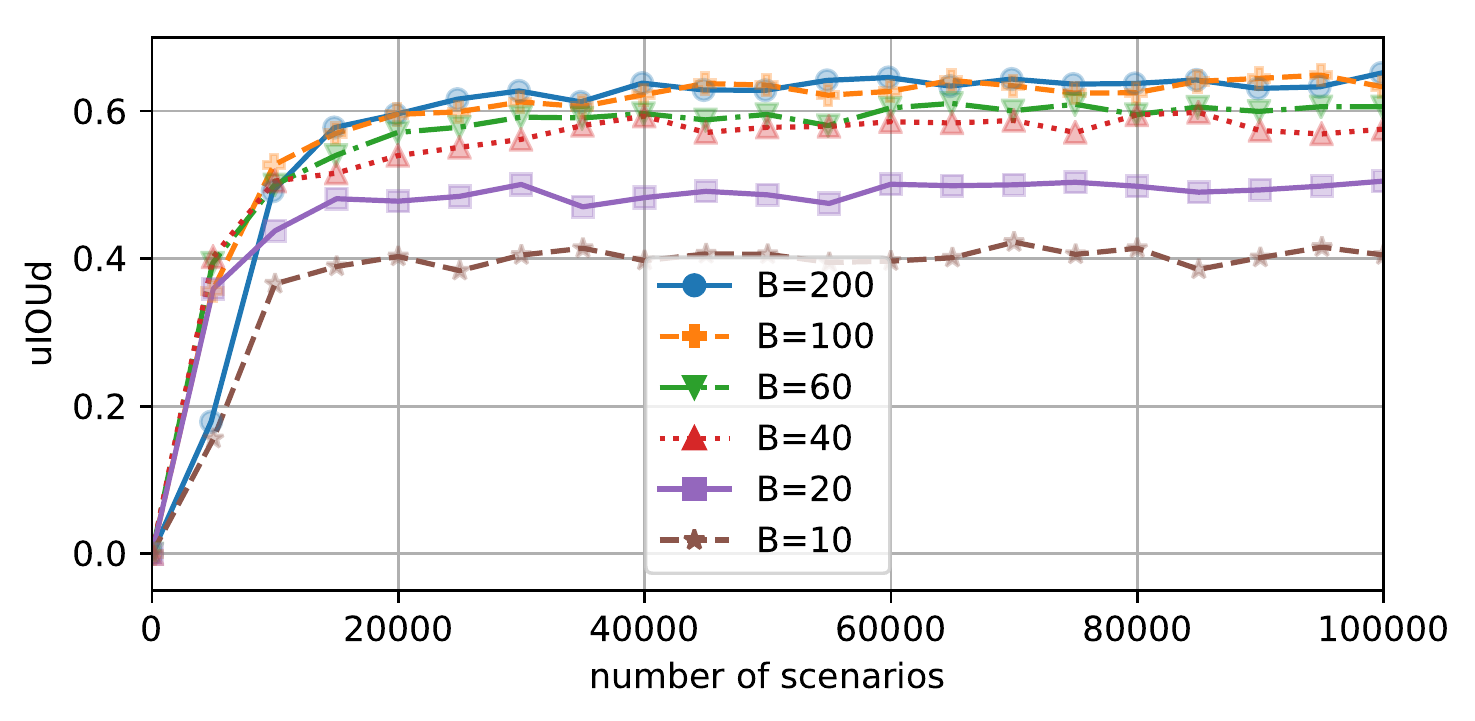}
        \caption{EOT minibatch size $B$}
        \label{fig:sensitivity_eotiters}
    \end{subfigure}
\caption{Adversarial strength over attack iterations, for various $\alpha$ values and EOT minibatch size.}
\label{fig:sensitivity_attack_rate}
\vspace{-0.2cm}
\end{figure}

We first optimized $\alpha$ for a fixed minibatch size of $B=20$.
As shown in Figure~\ref{fig:sensitivity_atkiters}, a step size of $\alpha=0.025$ attained the best end-performance, however $\alpha=0.075$ converged initially much faster.
This trade-off substantiates our empirical observations and suggests that the source texture initially needs to have most of its pixels \emph{broadly} perturbed to cause adversarial texture patterns to emerge, which would require drastic pixel changes with large perturbation sizes.
Subsequently, however, slight \emph{localized} pixel enhancements around \enquote{critical adversarial patterns} (see Section~\ref{sec:imitation}) steadily enhance the PAT's adversarial strength.
Thus, we recommend a practical schedule that starts with a large perturbation size of $\alpha=0.075$ for $500$ attack iterations, and then refines using a smaller step size of $\alpha=0.025$.

Next, using a single non-scheduled perturbation size of $\alpha=0.075$, we varied the EOT minibatch size $B$.
Note that, in Figure~\ref{fig:sensitivity_eotiters}, $\mu IOUd$ is plotted against the number of total EOT scenarios observed, i.e., $B \times I$.
These results show consistent performance trends that are proportional to $B \times I$, i.e. the total number of scenes seen by each PAT attack, rather than the number of attack iterations $I$ itself.
Also, beyond small values of $B \ge 20$ that lead to high-variance stochastic gradient updates, larger minibatch sizes result in similar and \emph{diminishing} amounts of improvement in both initial convergence speed and asymptotic adversarial strength.
Consequently, we chose $B = 20$ for the best trade-off between compute per attack iteration and convergence.

\subsection{Texture Attributes} \label{sec:appendix_exp_texture_resolution}

\begin{figure}[h]
    \centering
    \begin{subfigure}{1.0\linewidth}
        \centering
        \includegraphics[width=\textwidth]{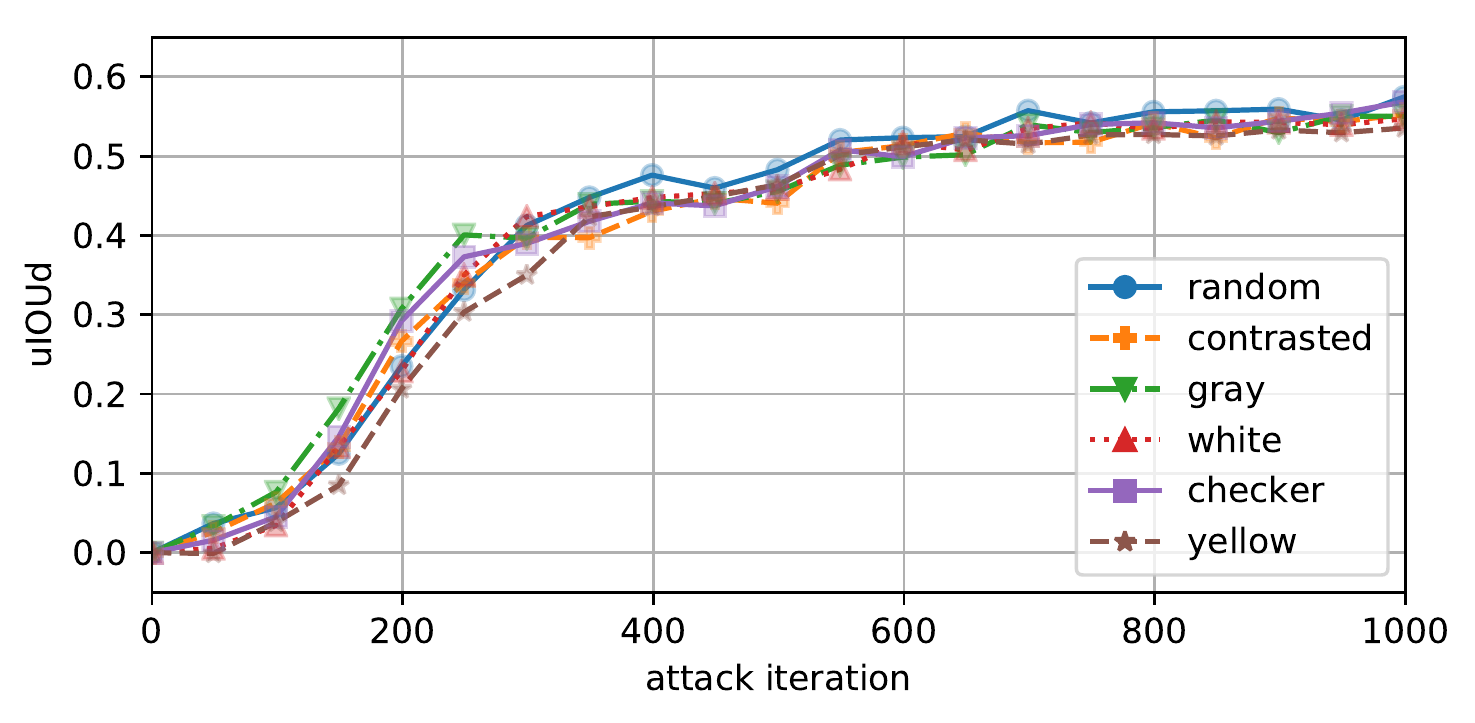}
        \caption{Initial textures}
        \label{fig:sensitiviy_eot_init_textures}
    \end{subfigure}
    \begin{subfigure}{1.0\linewidth}
        \centering
        \includegraphics[width=\textwidth]{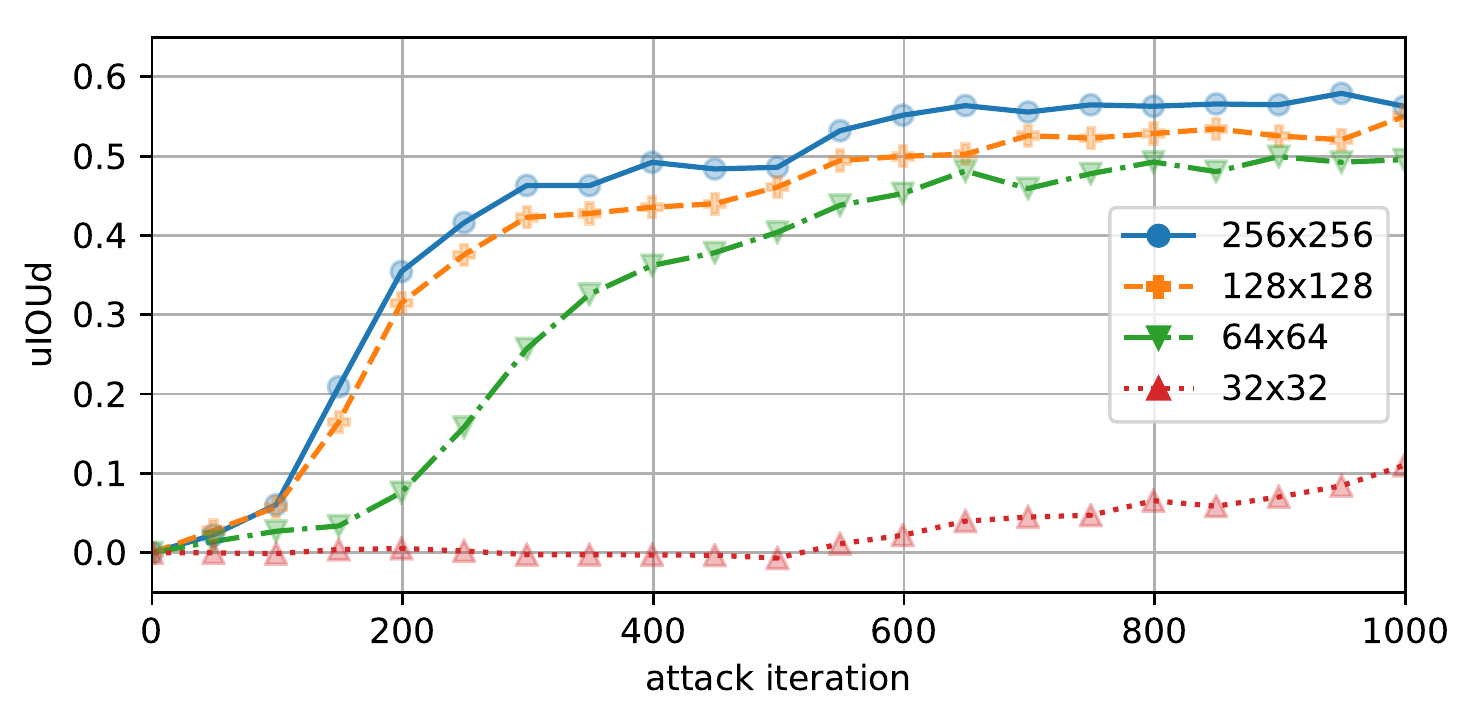}
        \caption{Texture sizes}
        \label{fig:sensitiviy_eot_texture_size}
    \end{subfigure}
\caption{Adversarial strength among various initial textures and texture sizes.}
\label{fig:sensitiviy_eot}
\end{figure}

Various related work made different recommendations on which source texture to use for best results.
In particular, suggestions included all-white and all-yellow~\cite{DBLP:journals/corr/abs-1812-10812}, and a \emph{random contrasted} checkerboard pattern alternating between uniform sampling of $[0,255]$ and $\{0,255\}$~\cite{zhang2018camou}.
We also tried an all-gray source pattern, as well as a per-pixel randomly-sampled source.

However, as shown in Figure~\ref{fig:sensitiviy_eot_init_textures}, we found that initializing the texture with different patterns did not result in significant changes in convergence nor performance. 

We also explored the effects of changing texture sizes and found that using a resolution of $32 \times 32$ lead to consistently poor results, while settings of $64 \times 64$, $128 \times 128$, and $256 \times 256$, yielded little differences in both initial convergence speed and asymptotic performance, as seen in Figure~\ref{fig:sensitiviy_eot_texture_size}.
We thus chose $128 \times 128$ to balance between having sufficient pixel capacity to accommodate the wide ranges of EOT conditions, and amount of computation to compute texture perturbations.
Still, we found it very important to be aware that our resolution choices are significantly affected by the viewing distances (see Appendix~\ref{sec:appendix_default_transformation}) and poster sizes used in our experiments.

\section{Ablation of EOT Conditioning Variables} \label{sec:appendix_eot_ablation}

In Section 6.3 of the main paper, we evaluated the effects of varying the ranges or choices for different EOT transformation variables, including background (\texttt{-bg}, \texttt{+bg}), target (\texttt{-target}, \texttt{+target}), lighting (\texttt{-light}, \texttt{+light}), poster size (\texttt{small poster}), camera pose (\texttt{-cam pose}, \texttt{+cam pose}), and target pose (\texttt{-target pose}, \texttt{+target pose}). Modified ranges to camera pose, target pose, and lighting are shown in Table~\ref{table:new_cam_pose},~\ref{table:new_target_pose}, and~\ref{table:new_lighting_diffuse}, respectively.
Also, variations for (\texttt{-bg}, \texttt{+bg}) and (\texttt{-target}, \texttt{+target}) are as follows:
\begin{itemize}
    \item \texttt{-bg}: use \texttt{playground} only;
    \item \texttt{+bg}: randomize among \texttt{school}, \texttt{forest}, \texttt{playground} and \texttt{cafe};
    \item \texttt{-target}: use \texttt{green person} only;
    \item \texttt{+target}: randomize among \texttt{green person}, \texttt{white person}, \texttt{t-shirt person}, \texttt{PR2} and \texttt{robonaut}.
\end{itemize}

\begin{table}[h]
\centering
\caption{PAT attack settings for \texttt{-cam pose} and \texttt{+cam pose}.}
\begin{tabular}{l|rr|rr}
\multirow{2}{*}{Transformation} & \multicolumn{2}{c|}{\texttt{-cam pose}} & \multicolumn{2}{c}{\texttt{+cam pose}} \\
{} & Min & Max & Min & Max \\
\hline
Initial x (m) & 0.0 & 0.0 & -2.0 & 2.0 \\
Initial y (m) & -8.5 & -8.5 & -16.5 & -5.5 \\
Initial z (m) & 1.2 & 1.2 & 0.4 & 2.2 \\
Initial roll (\degree) & 0.0 & 0.0 & -1.5 & 1.5 \\
Initial pitch (\degree) & 0.0 & 0.0 & -10.0 & 10.0 \\
Initial yaw (\degree) & 0.0 & 0.0 & -20.0 & 20.0 \\
\hline
$\Delta$x (m) & 0.0 & 0.0 & -0.15 & 0.15 \\
$\Delta$y (m) & 0.0 & 0.0 & -0.80 & 0.80 \\
$\Delta$z (m) & 0.0 & 0.0 & -0.15 & 0.15 \\
$\Delta$roll (\degree) & 0.0 & 0.0 & 0.0 & 0.0 \\
$\Delta$pitch (\degree) & 0.0 & 0.0 & -5.0 & 5.0 \\
$\Delta$yaw (\degree) & 0.0 & 0.0 & -5.0 & 5.0 \\
\hline
\end{tabular}
\label{table:new_cam_pose}
\end{table}

\begin{table}[h]
\centering
\caption{PAT attack settings for \texttt{-target pose} and \texttt{+target pose}.}
\begin{tabular}{l|rr|rr}
\multirow{2}{*}{Transformation} & \multicolumn{2}{c|}{\small \texttt{-target pose}} & \multicolumn{2}{c}{\small \texttt{+target pose}} \\
{} & Min & Max & Min & Max \\
\hline
Initial x (m) & 0.0 & 0.0 & -1.6 & 1.6 \\
Initial y (m) & -2.7 & -2.7 & -5.0 & -0.7 \\
Initial z (m) & 0.0 & 0.0 & 0.0 & 0.0 \\
Initial roll (\degree) & 0.0 & 0.0 & 0.0 & 0.0 \\
Initial pitch (\degree) & 0.0 & 0.0 & 0.0 & 0.0 \\
Initial yaw (\degree) & 90.0 & 90.0 & -90.0 & 270.0 \\
\hline
$\Delta$x (m) & 0.0 & 0.0 & -0.15 & 0.15 \\
$\Delta$y (m) & 0.0 & 0.0 & -0.15 & 0.15 \\
$\Delta$z (m) & 0.0 & 0.0 & 0.0 & 0.0 \\
$\Delta$roll (\degree) & 0.0 & 0.0 & 0.0 & 0.0 \\
$\Delta$pitch (\degree) & 0.0 & 0.0 & 0.0 & 0.0 \\
$\Delta$yaw (\degree) & 0.0 & 0.0 & -20.0 & 20.0 \\
\hline
\end{tabular}
\label{table:new_target_pose}
\end{table}

\begin{table}[ht]
\centering
\caption{PAT attack settings for \texttt{-light} and \texttt{+light}.}
\begin{tabular}{l|rr|rr}
\multirow{2}{*}{Diffuse Light Source} & \multicolumn{2}{c|}{\texttt{-light}} & \multicolumn{2}{c}{\texttt{+light}} \\
{} & Min & Max & Min & Max \\
\hline
Hue & 0.0 & 360.0 & 0.0 & 360.0 \\
Saturation & 0.0 & 0.0 & 0.0 & 0.7 \\
Value & 0.7 & 0.7 & 0.0 & 0.7 \\
\hline
\end{tabular}
\label{table:new_lighting_diffuse}
\end{table}

\begin{figure}[hb!]
    \centering
    \begin{subfigure}{1.0\linewidth}
        \centering
        \includegraphics[width=\textwidth]{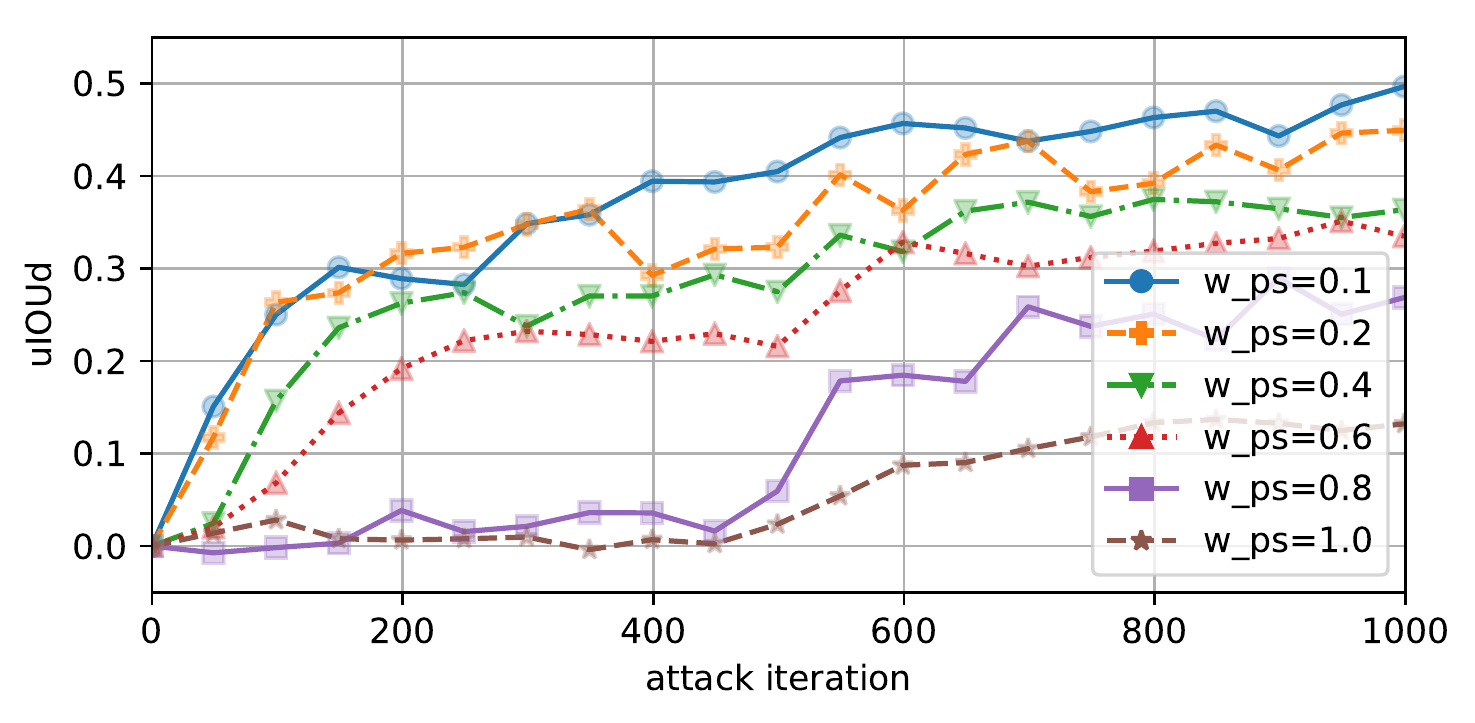}
        \caption{uIOUd}
        \label{fig:imitation_wperc_uIOUd}
    \end{subfigure}
    \begin{subfigure}{1.0\linewidth}
        \centering
        \includegraphics[width=\textwidth]{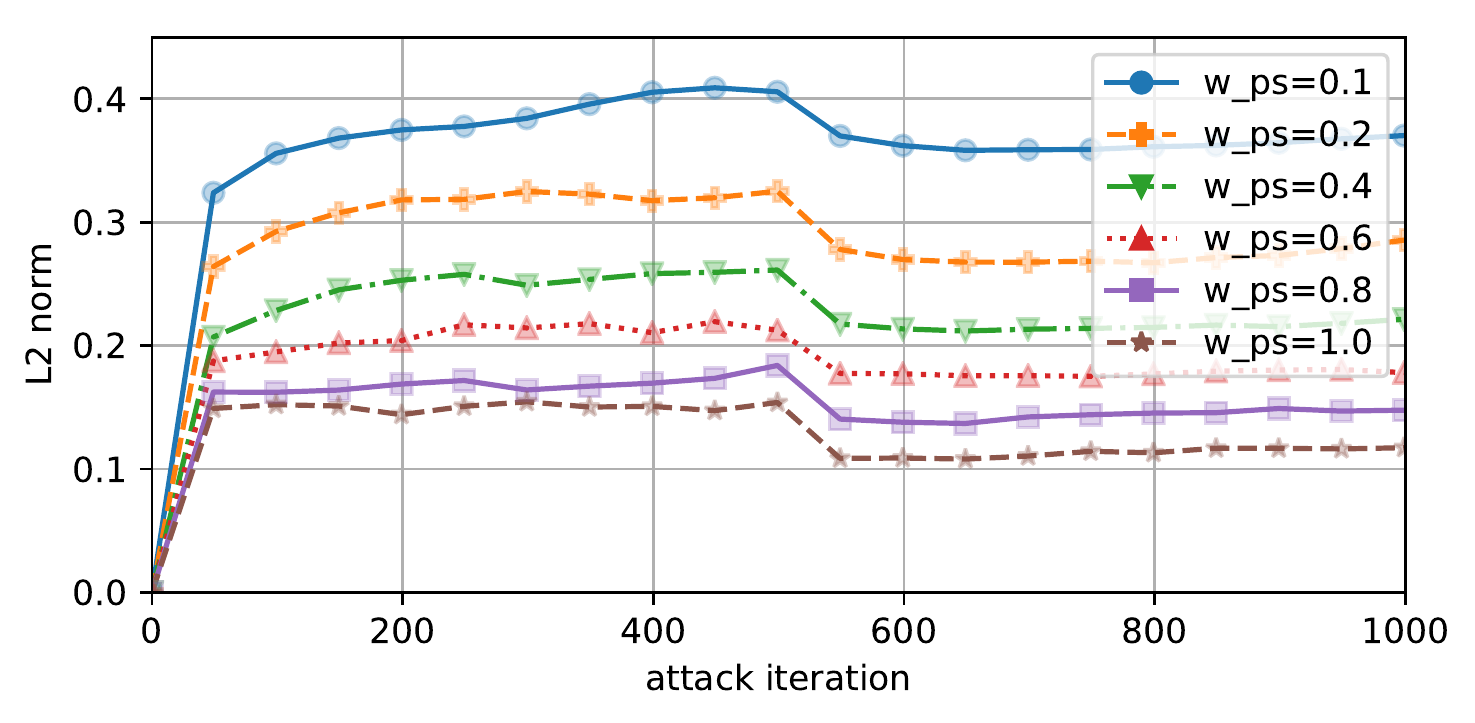}
        \caption{$L_2$-norm perceptual similarity}
        \label{fig:imitation_wperc_L2d}
    \end{subfigure}
\caption{Adversarial strength and perceptual similarity among various $w_{ps}$.}
\label{fig:imitation_wperc}
\end{figure}

\section{Imitation Attacks} \label{sec:appendix_imitation}

\begin{figure*}[ht]
\centering
\includegraphics[width=0.85\linewidth]{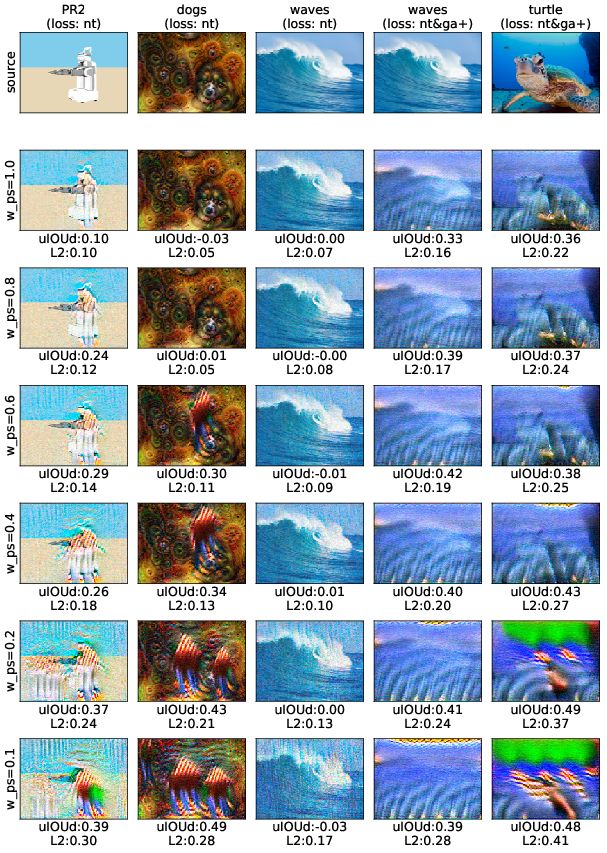}
\caption{PATs generated with various $w_{ps}$ values.}
\label{fig:pat_various_wps}
\end{figure*}

In Section~\ref{sec:imitation}, we set the value of $w_{ps}=0.6$. This value was determined based on an experiment where we studied the effect of changing $w_{ps}$ on the adversarial strength $\mu IOUd$ and perceptual similarity (as measured by the Euclidean $L_2$ distance to the source image in RGB colorspace).
Unsurprisingly, as seen in Figure~\ref{fig:imitation_wperc}, smaller values of $w_{ps}$ imposed fewer constraints and thus lead to faster attack convergence and better end-performance, while the inverse was true for larger values of $w_{ps}$.
We thus chose $w_{ps}=0.6$ after manually assessing which PATs had recognizable levels of perceptual similarity to their source images, as seen in Figure~\ref{fig:pat_various_wps}.

\begin{figure}[b]
    \centering
    \begin{subfigure}{1.0\linewidth}
        \centering
        \includegraphics[width=\textwidth]{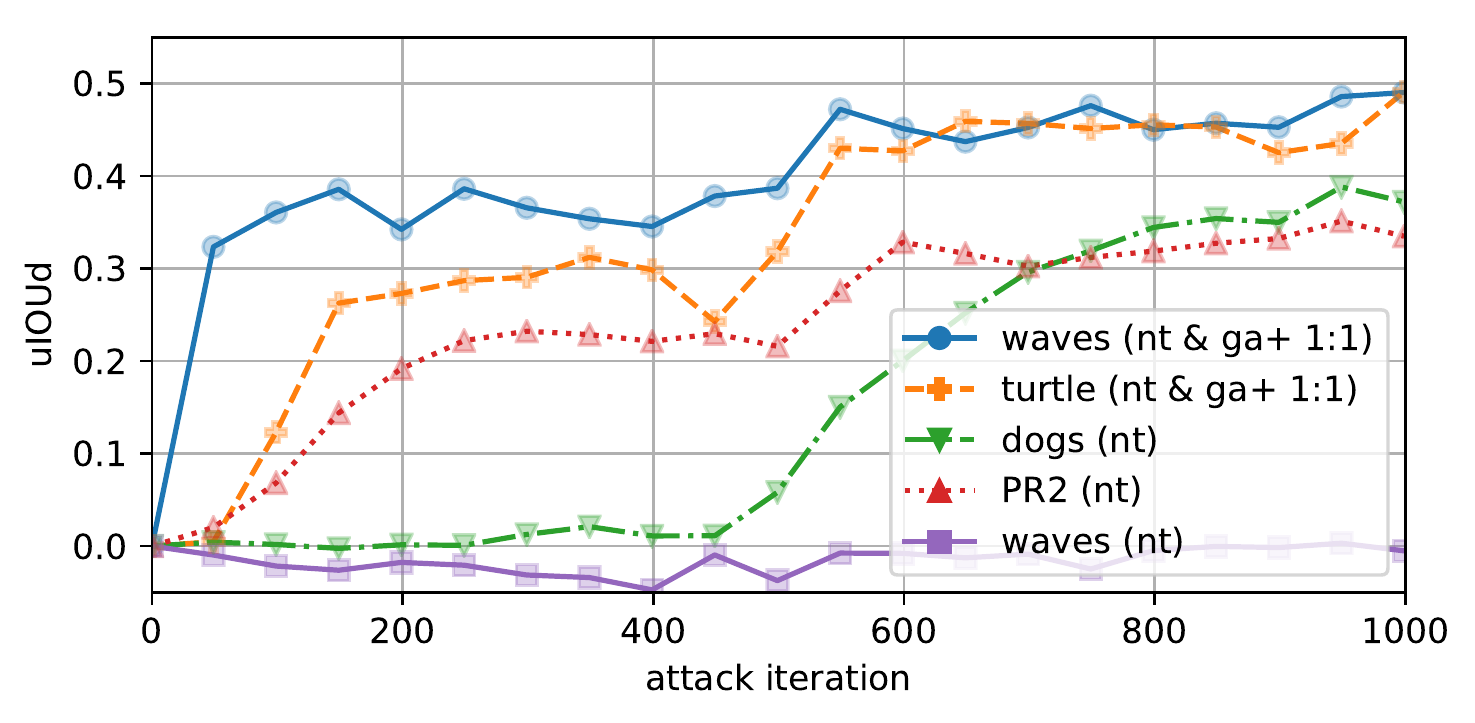}
        \caption{uIOUd}
        \label{fig:imitation_sources_uIOUd}
    \end{subfigure}
    \begin{subfigure}{1.0\linewidth}
        \centering
        \includegraphics[width=\textwidth]{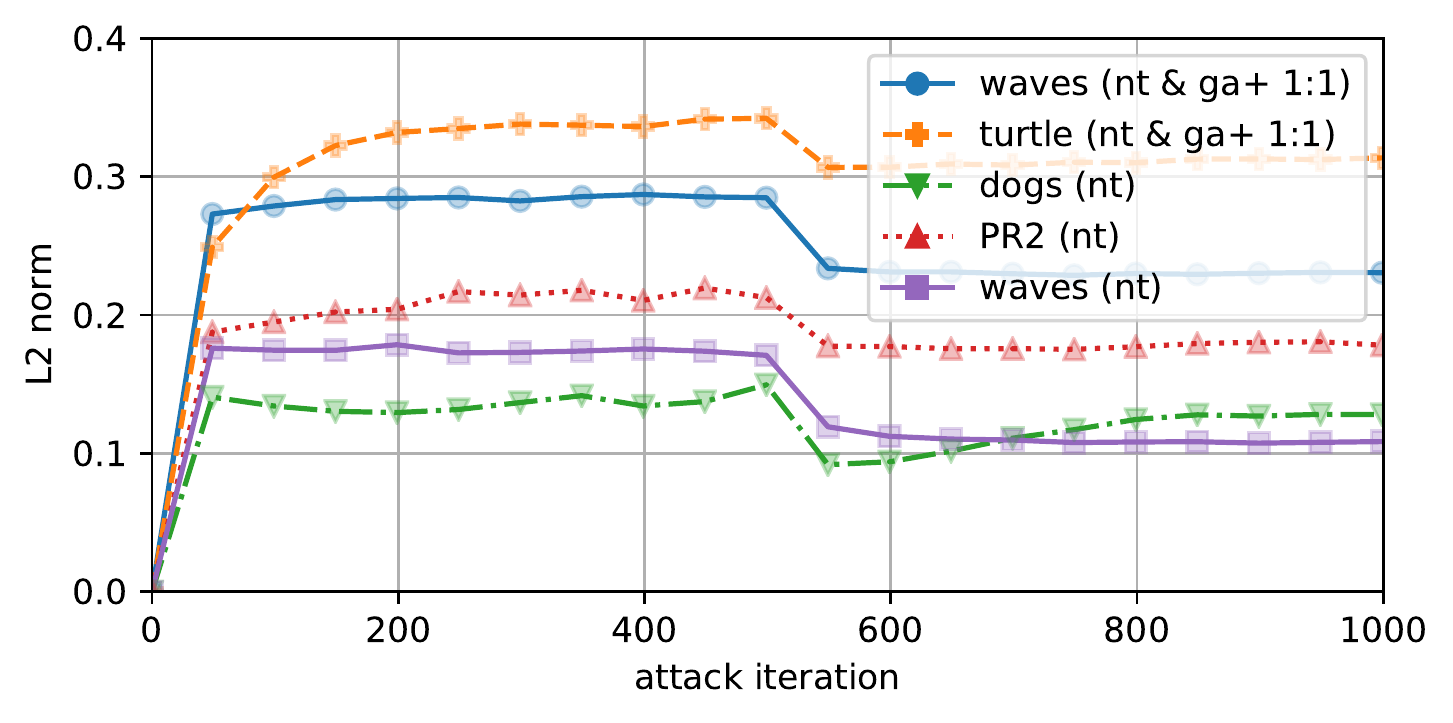}
        \caption{$L_2$-norm perceptual similarity}
        \label{fig:imitation_sources_L2d}
    \end{subfigure}
\caption{Adversarial strength and perceptual similarity among source textures shown in Figure 6 of main paper.}
\label{fig:imitation_sources}
\end{figure}

To substantiate Figure~\ref{fig:imitation_samples}, Figure~\ref{fig:imitation_sources} illustrates how $\mu IOUd$ and perceptual similarity metrics change over attack iterations.
As we can see, some specific combinations of initial posters and losses made the attack easier to converge.
For example, performing attacks using \texttt{waves} as initial texture with hybrid losses ($\mathcal{L}_{nt} \ \& \ \mathcal{L}_{ga+}$) resulted in strong adversaries, while using non-targeted loss alone did not.

\begin{figure*}[ht!]
    \centering
    \begin{subfigure}{0.49\linewidth}
        \centering
        \includegraphics[width=\textwidth]{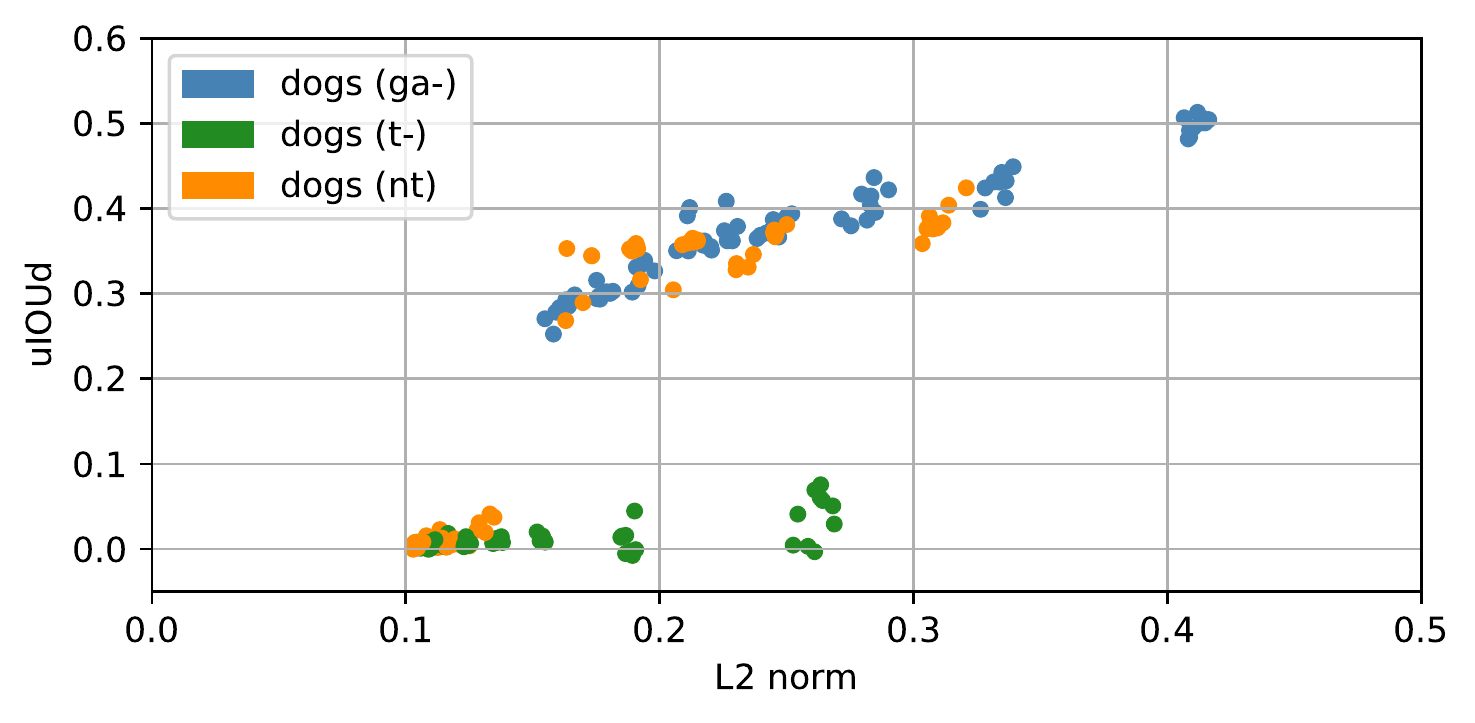}
        \caption{\texttt{dogs} at attack iteration 500}
        \label{fig:eval_imitation_dogs_500}
    \end{subfigure}
    \hbox{~}
    \begin{subfigure}{0.49\linewidth}
        \centering
        \includegraphics[width=\textwidth]{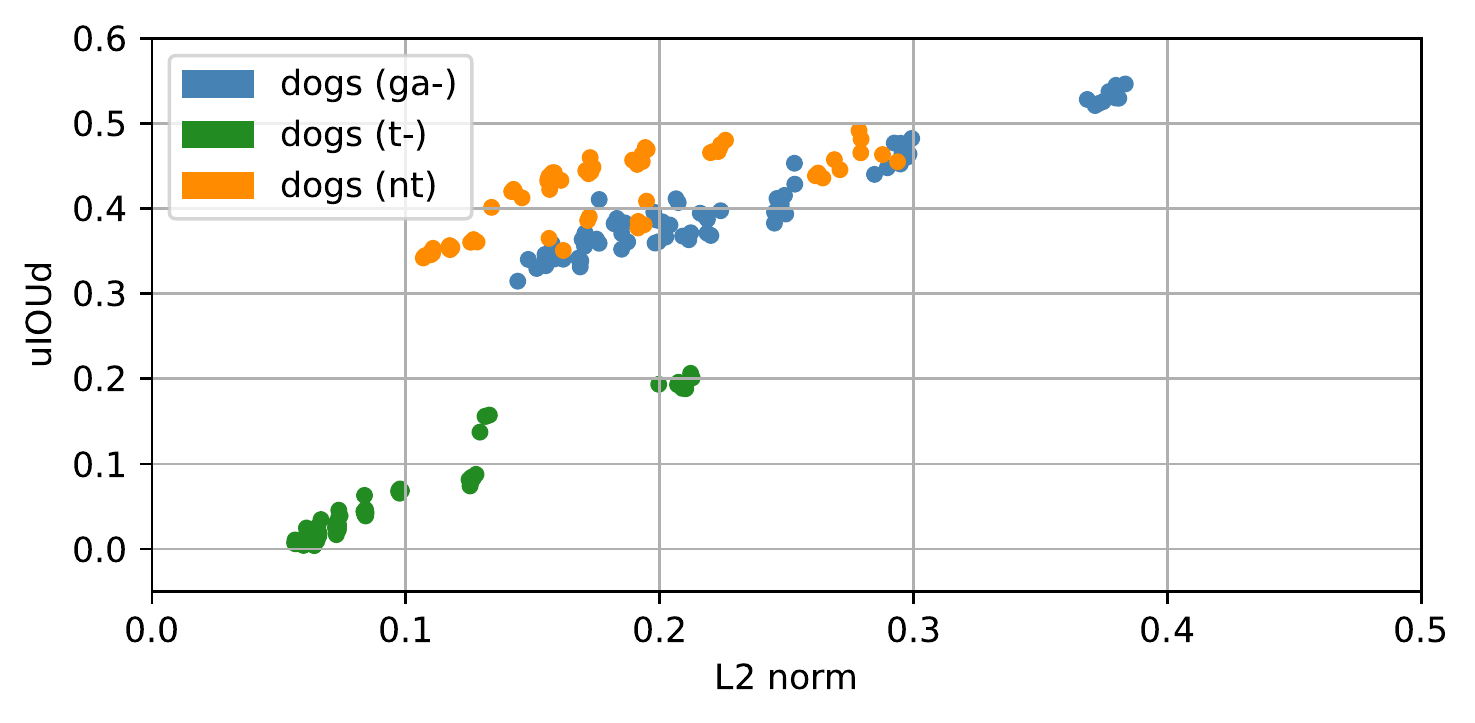}
        \caption{\texttt{dogs} at attack iteration 1000}
        \label{fig:eval_imitation_dogs_1000}
    \end{subfigure}
    \begin{subfigure}{0.49\linewidth}
        \centering
        \includegraphics[width=\textwidth]{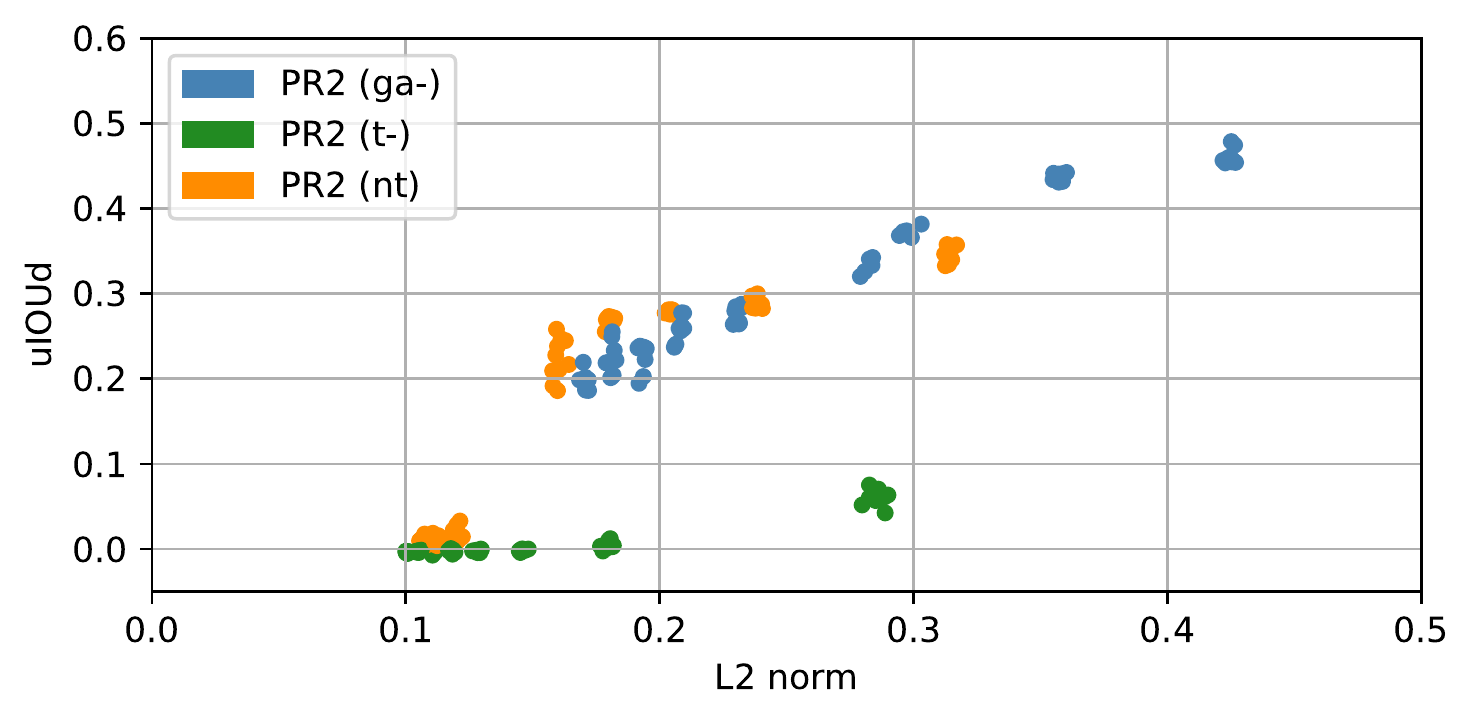}
        \caption{\texttt{PR2} at attack iteration 500}
        \label{fig:eval_imitation_pr2_500}
    \end{subfigure}
    \hbox{~}
    \begin{subfigure}{0.49\linewidth}
        \centering
        \includegraphics[width=\textwidth]{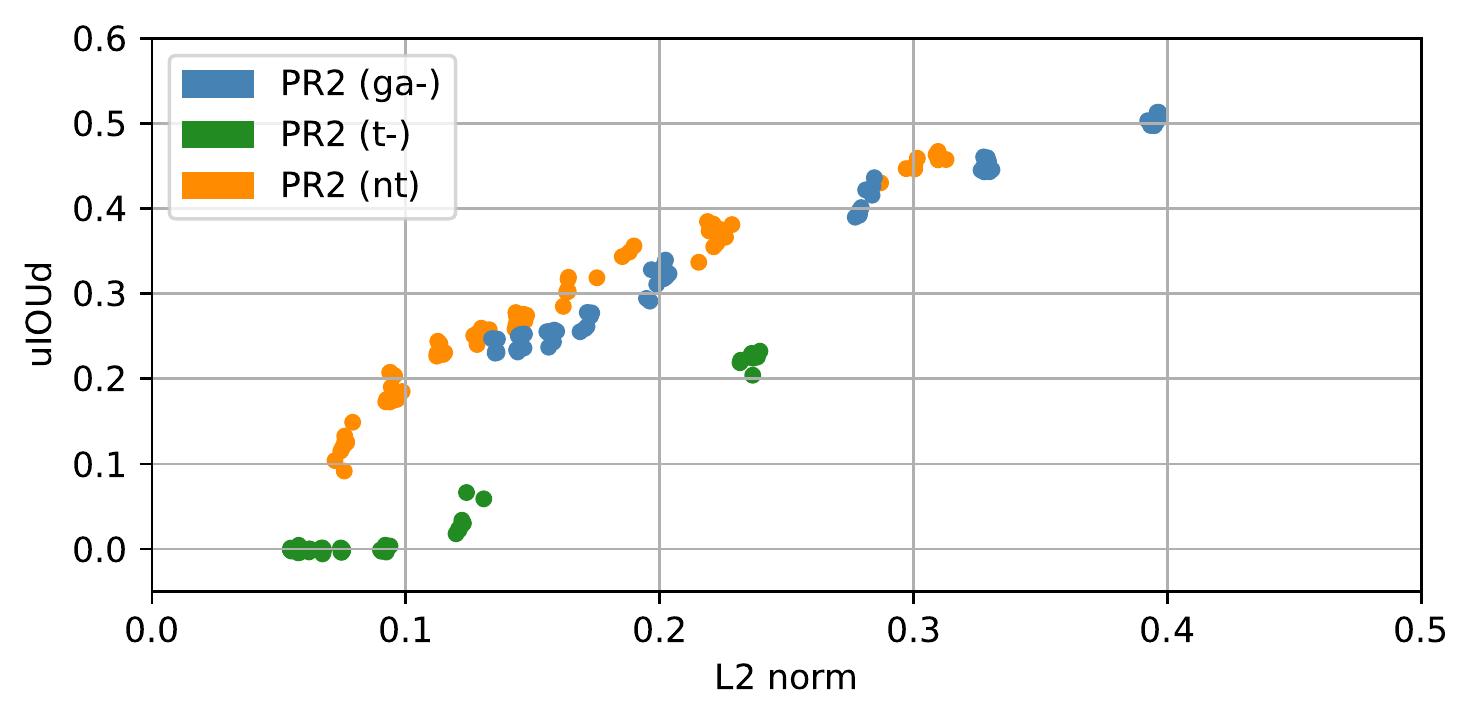}
        \caption{\texttt{PR2} at attack iteration 1000}
        \label{fig:eval_imitation_pr2_1000}
    \end{subfigure}
\caption{Performance of $\mathcal{L}_{nt}$, $\mathcal{L}_{ga-}$, and $\mathcal{L}_{t-}$ in imitating \texttt{dogs} and \texttt{PR2} textures for $w_{ps} \in [0.1:0.1:0.8]$.}
\label{fig:eval_imitation}
\end{figure*}

Figure~\ref{fig:eval_imitation} compares perceptual similarity ($L_2\,norm$) against adversarial strength ($\mu IOUd$) among $\mathcal{L}_{ga-}$, $\mathcal{L}_{nt}$, and $\mathcal{L}_{t-}$, and shows that, for a given threshold on $L_2\,norm$, $\mathcal{L}_{ga-}$ generally had better early convergence, but likely weakened adversarial strength after more attack iterations.

\begin{figure*}[ht]
    \centering
    \begin{subfigure}{1.0\linewidth}
        \centering
        \includegraphics[width=\textwidth]{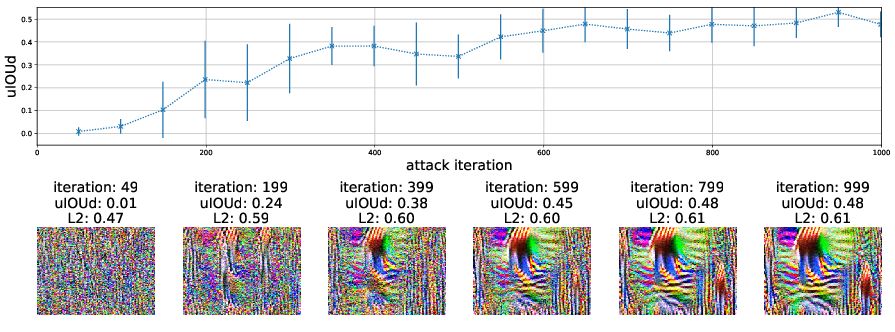}
        \caption{Non-imitation attack}
        \label{fig:cap_non_imitation}
    \end{subfigure}
    \begin{subfigure}{1.0\linewidth}
        \centering
        \includegraphics[width=\textwidth]{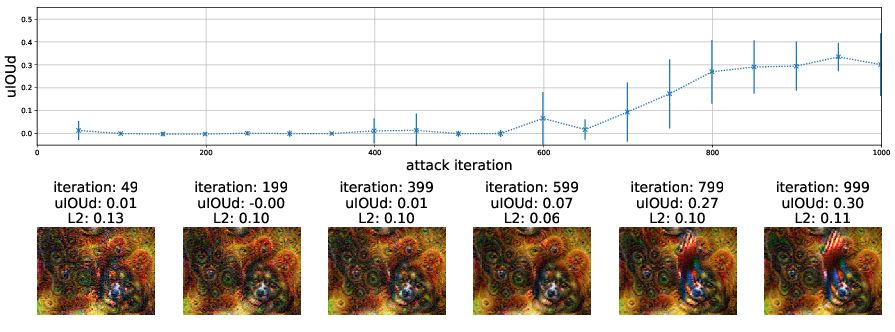}
        \caption{Imitation attack}
        \label{fig:cap_imitation}
    \end{subfigure}
\caption{The emergence of \enquote{critical adversarial patterns} for non-imitation and imitation attacks.}
\label{fig:cap_emergence}
\end{figure*}

Figure~\ref{fig:cap_emergence} illustrates the emergence of \enquote{critical adversarial patterns} that we discussed in Section 6.4 in the main paper.
In Figure~\ref{fig:cap_non_imitation}, the \emph{critical} dark striped pattern started to emerge at around iteration $400$, followed by the appearances of other nearby colorful patterns, which presumably were to drive predictions towards the central adversarial striped pattern.
In contrast, when we imposed a perceptual similarity loss during an imitation attack, only the dark striped pattern eventually emerged after significantly more attack iterations, as seen in Figure~\ref{fig:cap_imitation}.


\section{Transfer among tracking models}

\begin{figure}[ht]
    \centering
    \begin{subfigure}{1.0\linewidth}
        \centering
        \includegraphics[width=\textwidth]{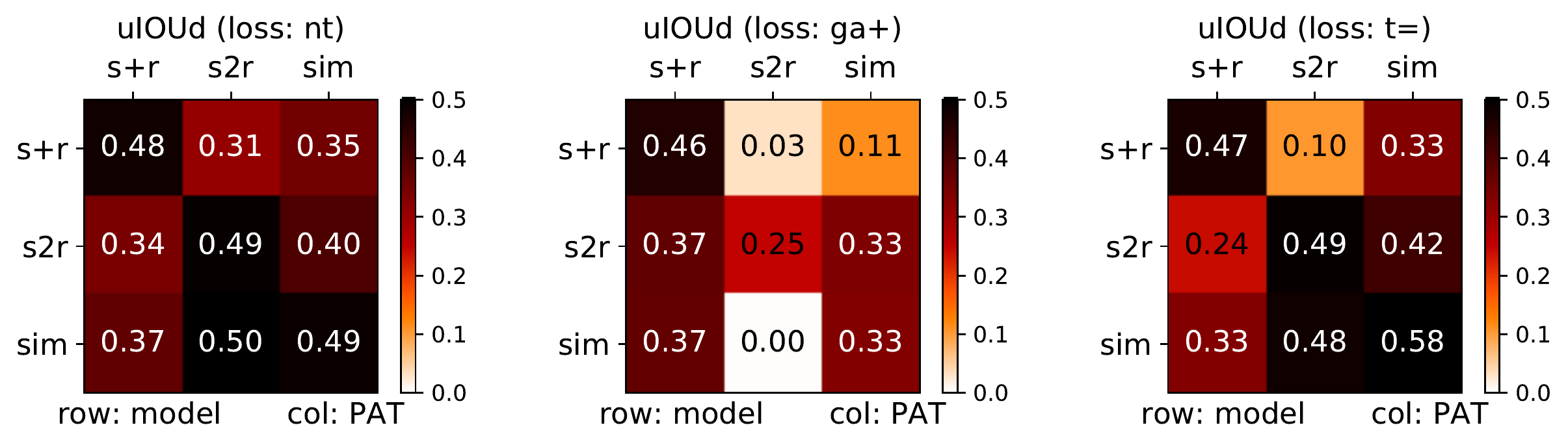}
        \caption{Sim\&real (\texttt{s+r}), sim-to-real (\texttt{s2r}), \texttt{sim}-only trained models}
        \label{fig:transfer_models}
    \end{subfigure}
    \begin{subfigure}{1.0\linewidth}
        \centering
        \includegraphics[width=\textwidth]{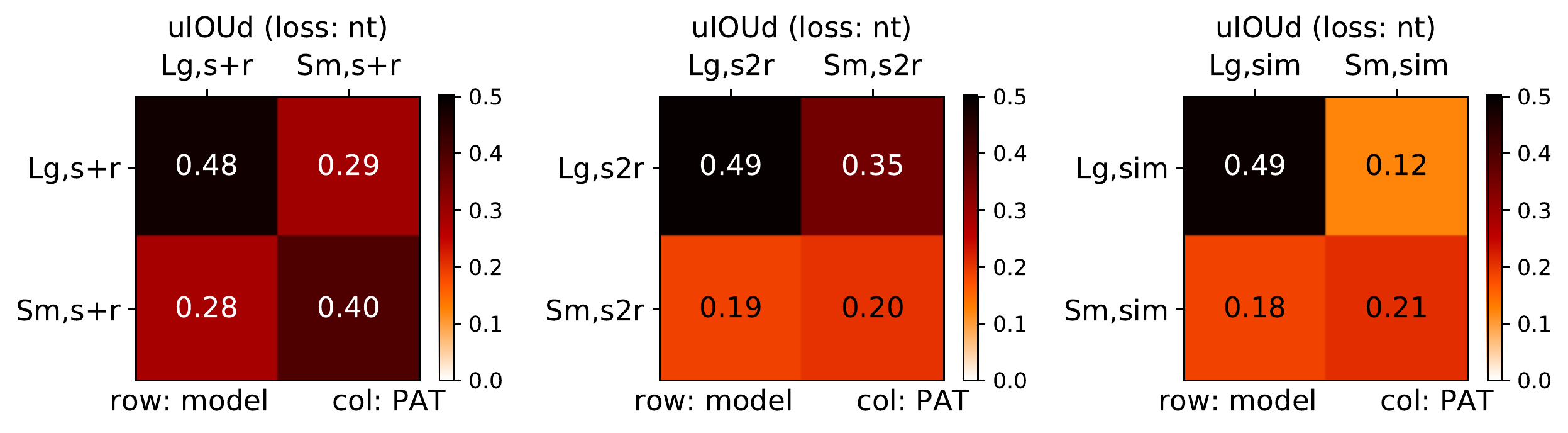}
        \caption{Models with default (\texttt{Lg}) and reduced (\texttt{Sm}) capacities}
        \label{fig:transfer_capacities}
    \end{subfigure}
\caption{Adversarial strength of generated PATs (columns) applied to different GOTURN tracking models (rows).}
\label{fig:transfer}
\end{figure}

We evaluated the transferability of PATs among different tracking models. 
When evaluating PATs on GOTURN models trained using different datasets, the off-diagonal results in Figure~\ref{fig:transfer_models} generally show that a decent-to-great amount of adversarial strength is still present.
Nevertheless, we see that the transferred efficacy of adversaries varied based on the tracker model and the loss used.
For instance, a \texttt{sim}-trained PAT optimized using $\mathcal{L}_{nt}$ and applied to the \texttt{s2r} GOTURN tracker is strongly adversarial, whereas a similar PAT optimized using $\mathcal{L}_{ga+}$ becomes completely inert.

Similarly, PATs preserved some of their adversarial strength when transferred between trackers with different capacities, as seen in Figure~\ref{fig:transfer_capacities}.
However, while all PATs applied to reduced-capacity models (\texttt{Sm}) affected GOTURN predictions, their $\mu IOUd$ values around $0.20$ do not reflect strong adversaries, thus indicating that it is more difficult to fool small-capacity GOTURN networks into consistently breaking away from their intended target.

\begin{figure*}[ht]
    \centering
    \begin{subfigure}{0.65\linewidth}
        \centering
        \includegraphics[width=\textwidth]{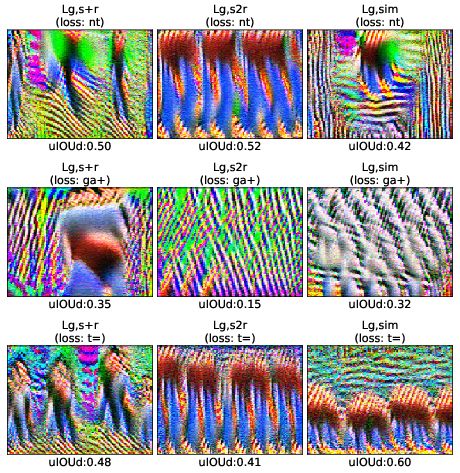}
        \caption{Different training datasets for GOTURN models}
        \label{fig:pat_xfermodel}
    \end{subfigure}
    \begin{subfigure}{0.65\linewidth}
        \centering
        \includegraphics[width=\textwidth]{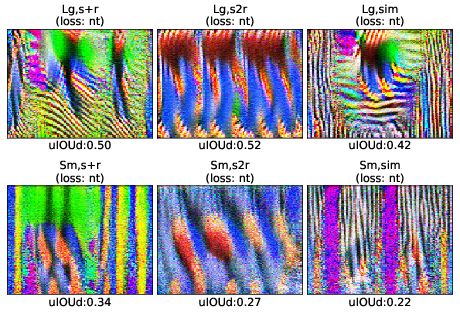}
        \caption{Different network capacities for GOTURN models}
        \label{fig:pat_xfercap}
    \end{subfigure}
    \caption{PATs used in the transferability among tracking models experiment.}
\label{fig:pat_transfer}
\end{figure*}

Figure~\ref{fig:pat_transfer} shows the PATs used in this experiment.
Generally, we observe similar adversarial patterns emerging from PAT Attacks on GOTURN models trained on different datasets, as well as different capacities, which explain why PATs transfer to a certain degree among different GOTURN trackers.
The sole exception is seen from the second row of Figure~\ref{fig:pat_xfermodel}, which reflected the fact that the adversarial loss $\mathcal{L}_{ga+}$ caused different patterns to emerge for different models, albeit with similar levels of competent adversarial strength.

\section{Demonstration of Sim-to-real Transfer} \label{sec:appendix_sim2real}

As discussed in Section~\ref{sec:sim2real}, we conducted test runs in real-world tracking and servoing conditions, and qualitatively verified the transferred adversarial strength of our synthetically-generated PATs, especially those containing \enquote{critical adversarial patterns}.
Many of these real-world runs are shown in the supplementary video.
Nevertheless, it is generally difficult to quantify performance consistently in the real world, due to tediousness and impracticality in labeling performance, controlling for repeated conditions, and dealing with practical complexities such as limited battery life and hardware failures.
Still, we segmented runs into video clips, and manually labeled them as either \texttt{strong}ly adversarial (i.e. where the tracker jumps onto the PAT and stays locked onto it even when momentarily obstructed), \texttt{weak}ly adversarial (i.e. where the tracker sometimes switches from the person to the PAT, and tends to latch back onto the person), or \texttt{fail}ure.

Looking at Table~\ref{table:physical_attack_performance}, we see that the tracker was quickly drawn to PATs when deployed on a stationary camera.
On the other hand, it was much harder to fool the person tracker when the drone was servoing the target.
Whether the PAT was displayed digitally on a monitor, or printed as an \texttt{A0} poster, we anecdotally observed that both of these materials displayed some amount of specular reflections.
These specularities changed as the camera moved around, and thus likely had altered the appearances of PATs during our servoing runs and rendered them inert.
Therefore, devising adversaries that are robust to specularities would be an exciting avenue for future research.

\begin{table}[ht]
\centering
\caption{Physical-world attack performance.}
\begin{tabular*}{0.41 \textwidth}{lrrr}
\hline
Runs & Strong & Weak & Fail \\
\hline
Stationary & 57 (71\%) & 13 (16\%) & 10 (13\%) \\
Servo & 6 (33\%) & 5 (28\%) & 7 (39\%) \\
\hline
\end{tabular*}
\label{table:physical_attack_performance}
\end{table}

\begin{figure*}[ht]
\centering
\includegraphics[width=1.0\linewidth]{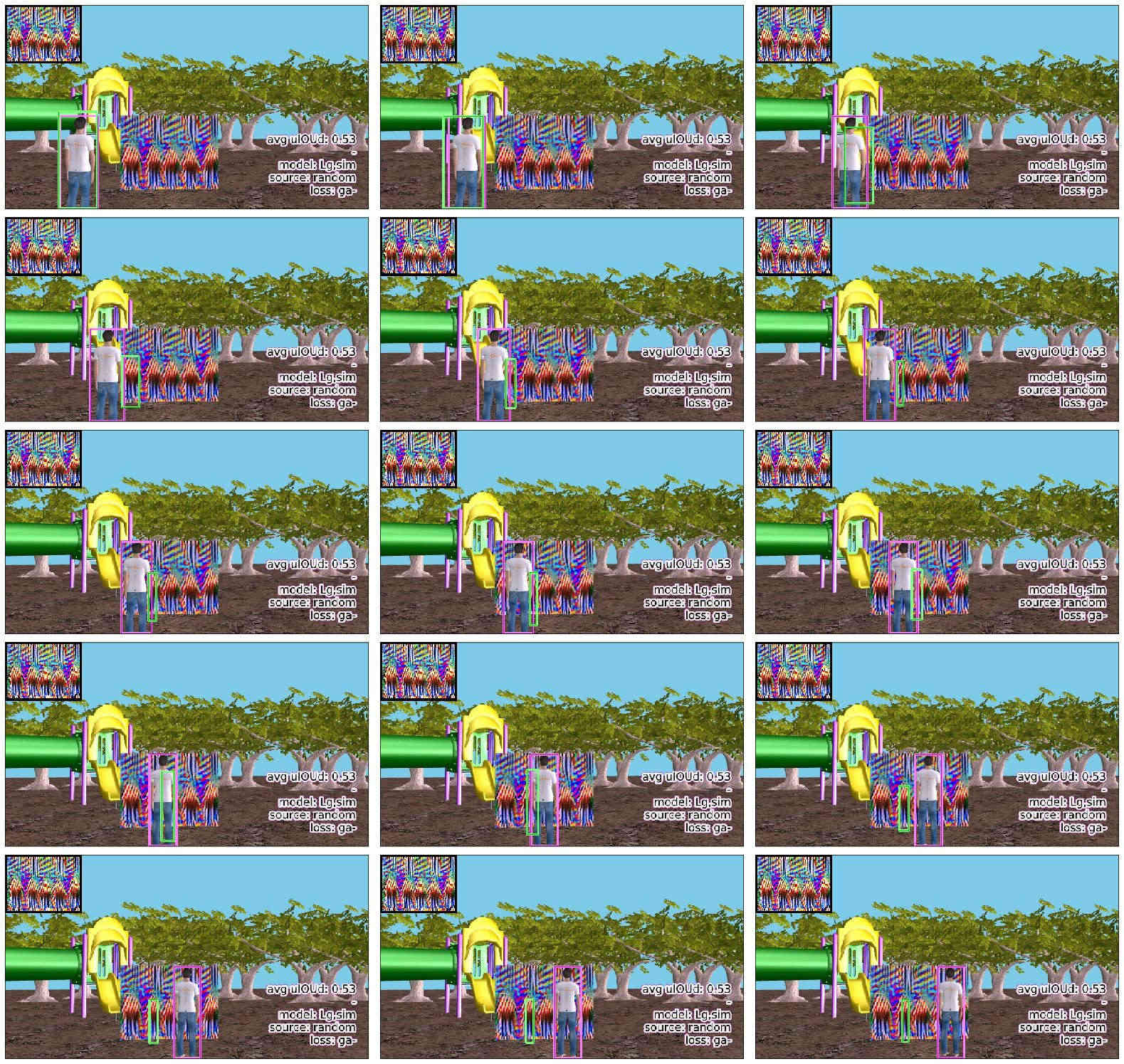}
\caption{PAT fools the tracker in simulation. Here, the purple bounding box represents the ground truth bounding box of the tracked object, while the green bounding box represents the tracker's prediction. Note that the sequence starts from the top-left frame to the bottom-right frame.}
\label{fig:sim_seq}
\end{figure*}

\begin{figure*}[ht]
\centering
\includegraphics[width=1.0\linewidth]{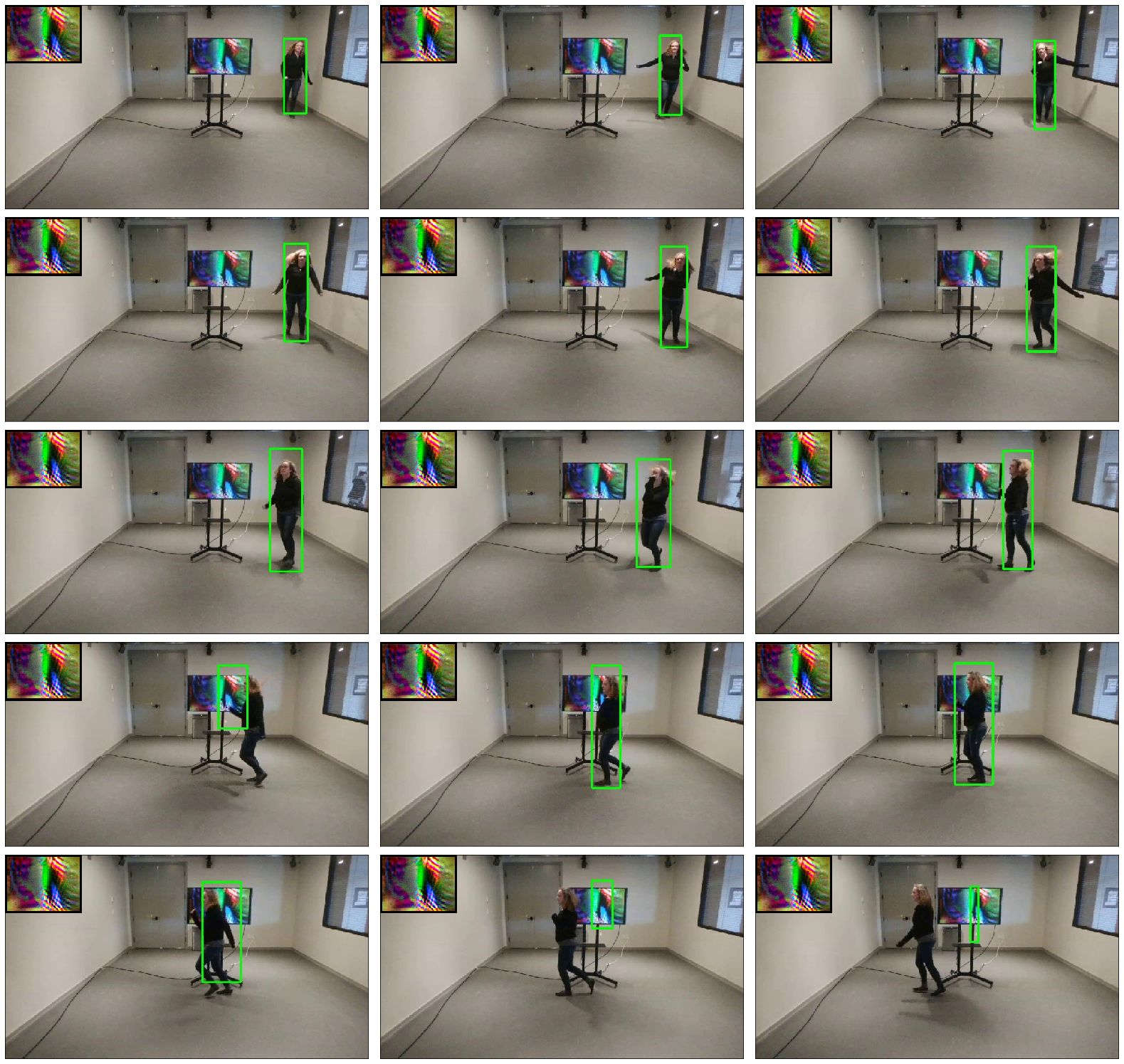}
\caption{PAT fools the tracker in the real world indoor setting, where the PAT is displayed on a TV. Note that the sequence starts from the top-left frame to the bottom-right frame.}
\label{fig:real_static_seq}
\end{figure*}

\begin{figure*}[ht]
\centering
\includegraphics[width=1.0\linewidth]{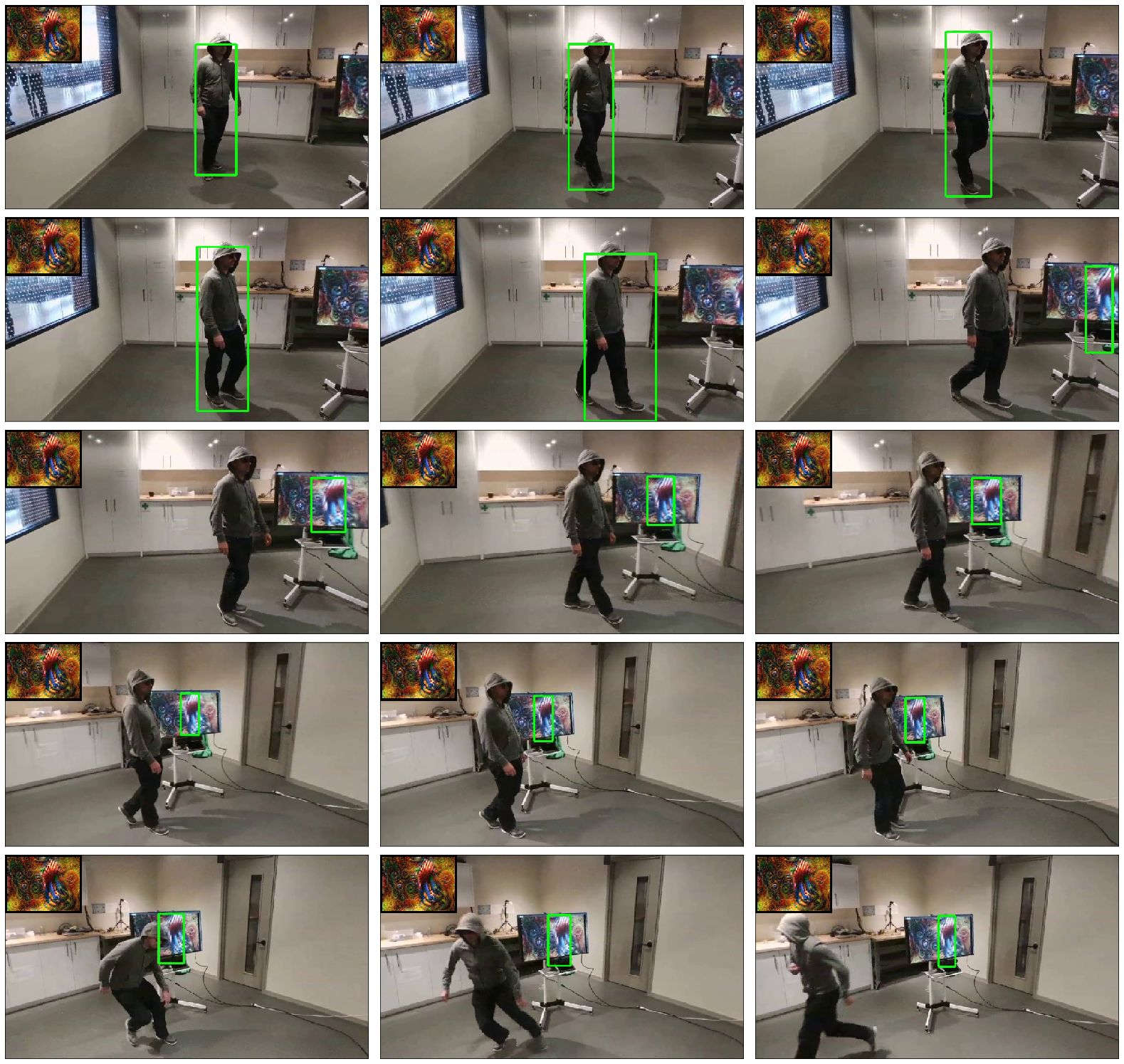}
\caption{PAT fools the tracker in the real world indoor setting during servoing run. Note that the sequence starts from the top-left frame to the bottom-right frame.}
\label{fig:real_servo_seq}
\end{figure*}

\begin{figure*}[ht]
\centering
\includegraphics[width=1.0\linewidth]{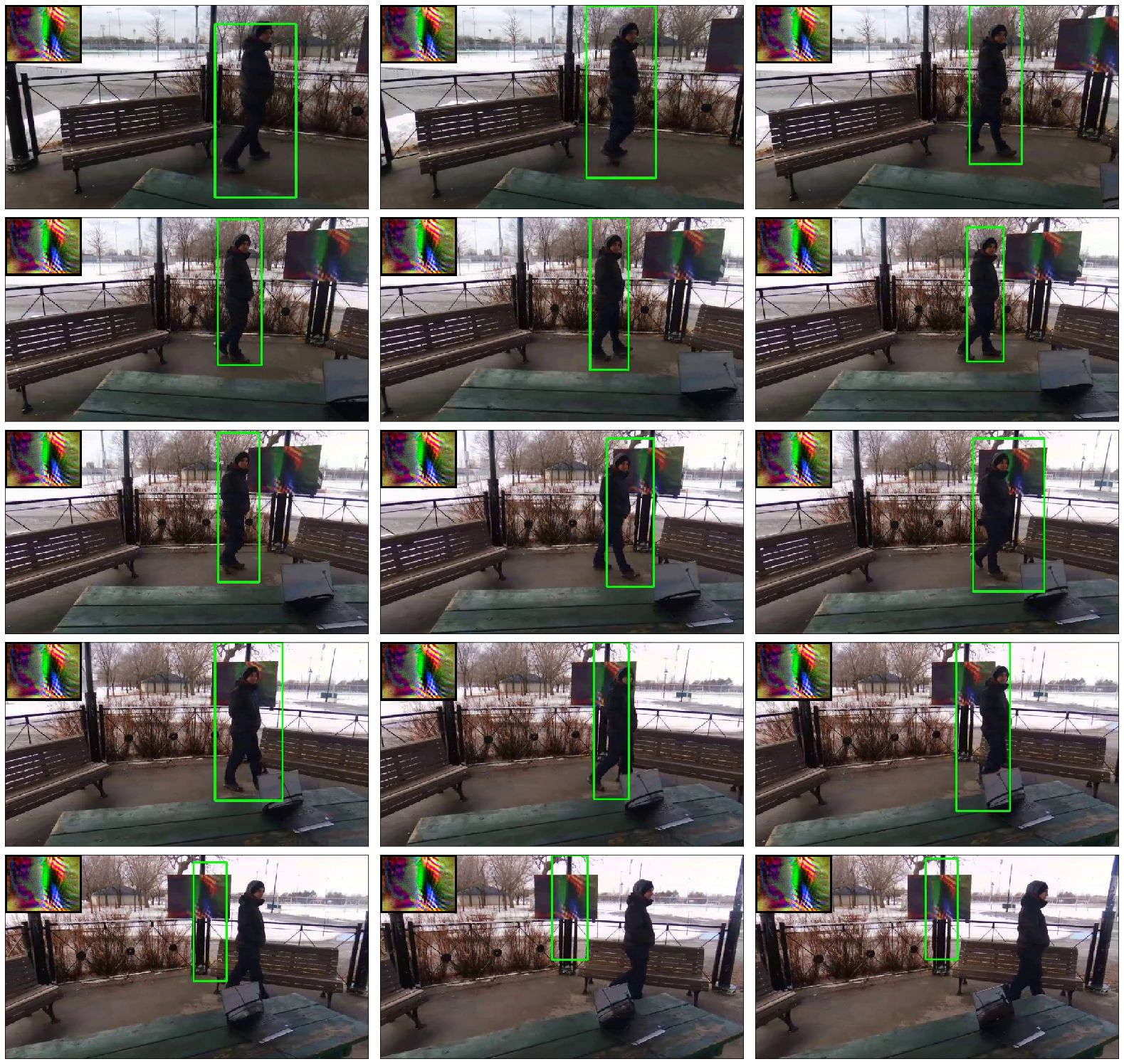}
\caption{PAT fools the tracker in the real world outdoor setting during servoing run, where the PAT is printed as a poster. Note that the sequence starts from the top-left frame to the bottom-right frame.}
\label{fig:real_servo_seq_poster}
\end{figure*}

\end{appendices}

\end{document}